\documentclass{article}

\usepackage[preprint]{corl_2026} 

\usepackage{enumitem} 
\usepackage{wrapfig}
\usepackage{graphicx}
\usepackage[normalem]{ulem}
\usepackage{caption}
\usepackage{booktabs}
\usepackage{tabularx}
\usepackage[most]{tcolorbox}
\usepackage{algorithm,algpseudocode}

\usepackage{subcaption}
\tcbset{
    boxsep=1pt,
    top=3pt,
    bottom=3pt,
    left=2pt,
    right=2pt,
}

\definecolor{takeawaygreen}{HTML}{568173}
\definecolor{takeawaybg}{HTML}{EEF3F1}

\newtcolorbox{keytakeaway}{
    enhanced,
    colback=takeawaybg,
    colframe=takeawaygreen,
    boxrule=0pt,
    arc=2mm,
    outer arc=2mm,
    left=4mm,
    right=4mm,
    top=1.5mm,
    bottom=1.5mm,
    borderline west={2.5pt}{0pt}{takeawaygreen},
}

\definecolor{accentgreen}{HTML}{568173}
\definecolor{questionbg}{HTML}{F8FAF9}

\newtcolorbox{questionbox}{
    enhanced,
    colback=questionbg,
    colframe=accentgreen,
    boxrule=0.8pt,
    arc=1.5mm,
    outer arc=1.5mm,
    left=3mm,
    right=3mm,
    top=1.5mm,
    bottom=1.5mm,
}

\definecolor{mypastelgreen}{HTML}{5B8E7D}  
\definecolor{mypastelred}{HTML}{BC4B51}  
\definecolor{mypastelorange}{HTML}{F4A259}  
\definecolor{myearth}{HTML}{A8986B}

\definecolor{mygreen}{rgb}{0.12, 0.54, 0.30}
\definecolor{myred}{rgb}{0.78, 0.33, 0.37}
\hypersetup{
    colorlinks=true,
    linkcolor=mypastelgreen,
    filecolor=magenta,      
    urlcolor=myearth,
    citecolor=mypastelgreen,
  pdftitle={Memory Anchors for Continual Robot Learning},
  pdfauthor={Maximilian Du; Zhanyi Sun; Chen Xu; Paarth Shah; Masha Itkina; Shuran Song},
}

\usepackage[nameinlink,capitalise]{cleveref}

\title{Memory Anchors for Continual Robot Learning}

\author{
  Maximilian Du\\
  Stanford University \\
  \texttt{maxjdu@stanford.edu} \\
  \And
    Zhanyi Sun\\
  Stanford University \\
  \texttt{zhanyis@stanford.edu} \\
  \And
    Chen Xu\\
  Toyota Research Institute \\
  \texttt{chen.xu@tri.global} \\
  \And
    Paarth Shah\\
  Toyota Research Institute \\
  \texttt{paarth.shah@tri.global} \\
  \And
    Masha Itkina\\
  Toyota Research Institute \\
  \texttt{masha.itkina@tri.global} \\
    \And
    Shuran Song\\
  Stanford University \\
  \texttt{shuran@stanford.edu} \\
}

\begin{document}
\maketitle


\begin{abstract}
    Robot policies deployed in the wild should have the capability to continually learn new tasks without forgetting existing behaviors. A common approach to combat such catastrophic forgetting is to train on new task data with a replay buffer of previously learned task data. Although this buffer is commonly sampled randomly from all prior experiences, we show that a small set of these experiences contributes greatly in anchoring past performance. We call these experiences \textit{Memory Anchors}. 
    We identify Memory Anchors in regions where representations of new-task observations collapse onto those of old-task observations even though the tasks require conflicting actions,
    like when a familiar object must be manipulated in a new way. Rehearsing old data in this region plays a key role in preventing destructive overwriting of past task knowledge, serving as this critical Memory Anchor role. Excluding only 10\% Memory Anchors before sampling the buffer leads to more than a 4.5x increase in catastrophic forgetting on the LIBERO benchmark suites. Conversely, enriching the replay buffer with Memory Anchors can decrease high-conflict task forgetting by $63$\% and enables successful continual learning of two task sequences on a real robot. 
\end{abstract}

\newcommand{\methodname}{\textsc{Anchor}ER}
\newcommand{\basemethod}{\texttt{RandER}}
\newcommand{\naivemethod}{\texttt{FixedER}}
\newcommand{\shuran}[1]{\textcolor{blue}{Shuran:#1}}
\newcommand{\chen}[1]{\textcolor{cyan}{Chen:#1}}
\newcommand{\zhanyi}[1]{{\color{purple}Zhanyi: #1}}
\section{Introduction}

As learned robot policies become increasingly capable, their continuous deployment raises the challenge of acquiring new capabilities without losing existing behaviors. New tasks are rarely independent of those already present in the policy: they may share observational similarities while requiring conflicting behaviors. For example, a robot opening a jar clockwise must both draw on and distinguish this new behavior from its experience opening jars counter-clockwise. 
Capturing these task relationships by balancing commonalities and contradictions is therefore essential for successful continual learning (CL). Failing to do so can lead to catastrophic forgetting: large drops in past task performance after learning the new task \cite{mccloskey1989catastrophic, french1999catastrophic}.

Existing continual learning approaches tackle catastrophic forgetting by regularizing new task learning with past knowledge~\cite{yang_continual_2026,verwimp_continual_2024}. A popular approach is Experience Replay (ER) \cite{chaudhry_tiny_2019}, which maintains a replay buffer of past data that is trained jointly with new task data, encouraging the policy to find solutions that keep both old and new task losses low.
ER is appealing due to its simplicity and agnosticism to architectures and training techniques \cite{liu_libero_2023, chaudhry_tiny_2019}. 

The benefits of ER generally carry over to robotic continual learning \cite{liu_libero_2023}, but in this work, we claim that their \textit{underlying mechanism} is more nuanced. We discover that a randomly-sampled ER buffer preserves some tasks easily during new task training, while other tasks degrade drastically. The affected tasks have conflict regions with the new task: high observation representation overlap (input) with very different required actions (output). Therefore, \uline{the small set of memories that regularize these conflicts play a disproportionately large role in anchoring past task performance.} We call this set: \textbf{Memory Anchors}. 


To test our hypothesis, we propose a three-step Memory Anchor extraction process.
First, we find the region of new task data that overlaps with old task data in the policy's latent space. 
Second, from this overlap, we extract the new data with the highest \textit{action disagreement}: the difference between the policy's \textit{predicted} actions and ground truth actions. Third, we use the policy's latent space to retrieve Memory Anchors: the old task data most similar to this conflict set. Using this method, we study how Memory Anchors influence catastrophic forgetting and also propose enriching their presence to improve past task retention.
Concretely, our contributions are as follows: 

\begin{itemize}[leftmargin=*, itemsep=0.5em, topsep=0.2em, parsep=0pt, partopsep=0pt]
 \item \textbf{Memory Anchors Analysis for Past Task Retention:} We posit that some data in the ER buffer contributes disproportionately to past task retention for robot continual learning. We extract these critical task-anchoring experiences (Memory Anchors) using the policy's predictions and representation space (\S \ref{sec:findingmemoryanchors}). We demonstrate the impact of these Memory Anchors by both reducing and enriching their presence in the ER buffer (\S \ref{sec:experiments}).

 \item \textbf{\methodname{} Algorithm for Reducing Catastrophic Forgetting:} 
 We propose \methodname{}, a method that enriches ER buffers with additional Memory Anchors to reduce catastrophic forgetting, especially for tasks that conflict when trained in sequence (\S \ref{sec:addanchors}).

 \item \textbf{Challenging Real-Robot Study:} Continual learning in robotics has mostly been studied in simulation benchmarks \cite{liu_libero_2023, liu2026pretrained}. From our Memory Anchor findings, we claim that \textit{homogeneous tasks} (tasks that are observationally similar but conflict in action) are the most difficult to learn sequentially. We design real robot task suites around this finding and validate \methodname{} on it (\S \ref{sec:realrobot}). 
\end{itemize} 

\begin{figure}[t]
  \centering
  \includegraphics[width=0.99\textwidth]{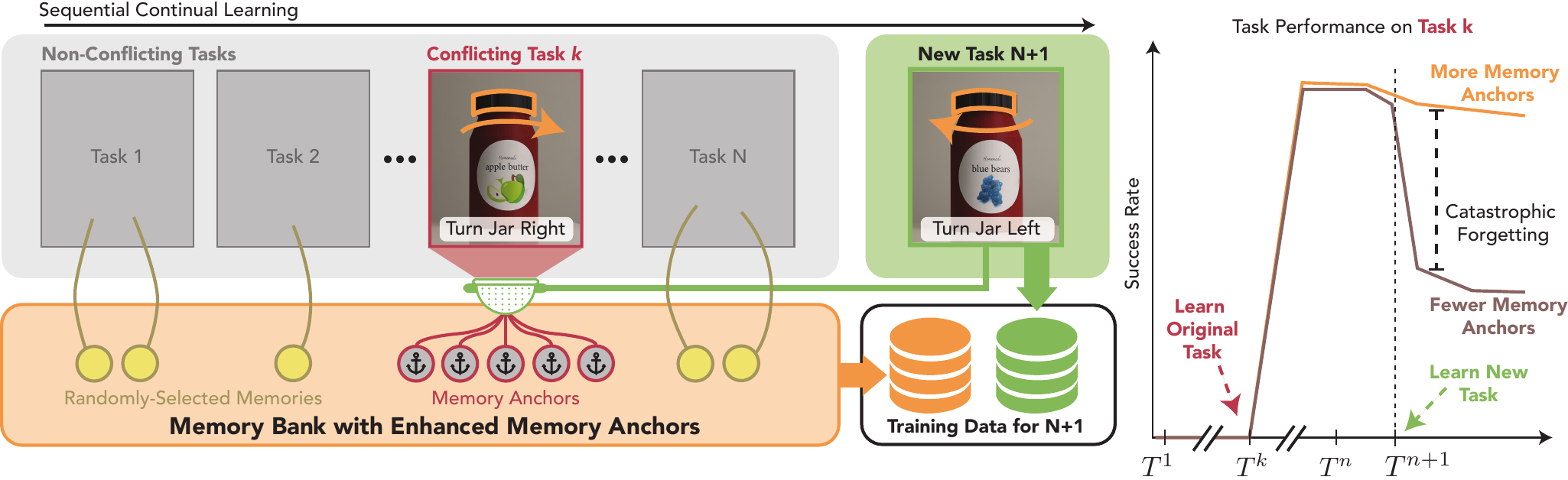}
\textbf{}  \caption{\textbf{Memory Anchors}. New tasks may be visually similar to old tasks while requiring different, sometimes even opposite actions, causing catastrophic forgetting (Right). We claim that a small proportion of old data plays a large role in anchoring past task performance. While these \textit{Memory Anchors} may appear naturally in randomly selected memories, their concentration is especially important with conflicting tasks. We propose a method for finding Memory Anchors and supplementing the ER buffer with them to reduce catastrophic forgetting for these worst-case scenarios (Left).}
  \label{fig:pullfig}
    \vspace{-7mm}
\end{figure}

On the LIBERO benchmark \cite{liu_libero_2023}, we find that reducing access to just 10\% of the best Memory Anchors leads to a $4.5$x increase in catastrophic forgetting on the same ER buffer size (\S \ref{sec:removeanchors}). Furthermore, reserving 10\% of an ER buffer for Memory Anchors (\methodname{}) leads to a $63$\% drop in catastrophic forgetting for the most conflicting task pairings (\S \ref{sec:addanchors}). The benefits of \methodname{} transfer to a real robot continual learning task (\S \ref{sec:realrobot}), where it improved the total task success by $1.7$x over a random ER buffer. Videos and additional visualizations can be found on our website: \href{https://robot-adaptation.github.io/MemoryAnchors}{robot-adaptation.github.io/MemoryAnchors}.

\section{Related Works}

\textbf{Continual Learning Methods:} Learning tasks in sequence yields two challenges that would not appear if the same tasks were trained all at once: \textit{catastrophic forgetting}, where adapting to new data overwrites knowledge of earlier tasks~\cite{mccloskey1989catastrophic,french1999catastrophic} and \textit{loss of plasticity}, where the model gradually loses its ability to learn new tasks~\cite{dohare2024loss}. Effective continual learning methods balance backwards transfer (stability) with forward transfer (plasticity)~\cite{de2021continual,lopez2017gradient}. Approaches to this challenge fall in three general families: regularization, architectures, and replay buffers \cite{yang_continual_2026}. Regularization approaches penalize changes to parameters deemed important for prior tasks~\cite{kirkpatrick2017overcoming,zenke2017continual,aljundi2018memory,yu2020gradient}. Architectural approaches dedicate separate parameters or subnetworks to
each task~\cite{rusu2016progressive,mallya2018packnet}. Replay approaches maintain a small buffer of past examples and sample them during new-task training~\cite{lopez2017gradient,chaudhry2018efficient,rebuffi2017icarl,riemer2018learning,buzzega2020dark, chaudhry_tiny_2019}, including the ER algorithm~\cite{chaudhry_tiny_2019,robins1995catastrophic}. Our work expands on replay buffer approaches and adapts claims of ER effectiveness~\cite{chaudhry_tiny_2019} to the robot task setting.

\textbf{Continual Learning in Robotics:}
In the robotics setting, continual learning must contend with further complications due to continuous action spaces, as well as diverse scenes and tasks~\cite{thrun1995lifelong,lesort2020continual,liu_libero_2023}. 
These complications also provide unique opportunities, including
reusing skills through evolving non-parametric knowledge
spaces~\cite{meng2025preserving,wu2025continually}, unsupervised skill
segmentation~\cite{wan2024lotus}, and expandable skill codebooks~\cite{xu2025speci}.
A second family of methods finetunes pretrained policies through task-specific adapters~\cite{liu2024tail,zhu2025efficient}, autonomous adapter
expansion~\cite{romer2026clare}, progressive expert
libraries~\cite{lei2025dynamic}, optimized finetuning
recipes~\cite{kim2025fine}, weight-space
averaging~\cite{yadav2025robust}, on-policy
RL~\cite{hu2026simple}, or iterative
self-improvement~\cite{bousmalis2023robocat}.
However, despite these diverse approaches, ER~\cite{chaudhry_tiny_2019} is still widely adopted for continual robot learning~\cite{xie2022lifelong,liu_libero_2023}, with recent work showing that even simple replay suffices to largely eliminate forgetting in pretrained VLAs~\cite{liu2026pretrained, zhu_can_2026}. Yet in these works, replay buffers are populated uniformly. Our work explores how inter-task relationships provide opportunities for a better ER sampling algorithm through Memory Anchors. We also construct a real-world example of continual learning that challenges a uniformly-sampled buffer (\S \ref{sec:realrobot}).

\textbf{Data Retrieval for Continual Learning:} Iterative task learning and refinement is sensitive to the training data. On-policy data has been shown to inherently resist forgetting compared to supervised finetuning on a fixed dataset ~\cite{shenfeld2025rl, hu2026simple}. If on-policy data is not available, ER is another way of reducing the data gap between existing and new tasks~\cite{robins1995catastrophic, prabhu2020gdumb}. Initial works stored small random subset of past data~\cite{chaudhry_tiny_2019,lopez2017gradient,rebuffi2017icarl,buzzega2020dark}, but subsequent works improved the buffer and its sampling strategies. Some store samples that maximize gradient diversity or approximate the full-data gradient~\cite{aljundi2019gradient,tiwari2022gcr}, while others retrieve the samples whose loss would spike under the
incoming update~\cite{aljundi2019online}. HAL synthesizes per-class anchors that maximize a hindsight estimate of forgetting, then regularizes predictions on
them~\cite{chaudhry2021using}. Our Memory Anchors operate on a similar principle, adapted for continuous visuomotor policies. Instead of generating anchors, we take inspiration from policy learning data retrieval approaches~\cite{memmel2025strap,lin2024flowretrieval,kumar2025collage} and select real data at the start of every task using state similarity and action disagreement (\S \ref{sec:findingmemoryanchors}).  
\newcommand{\der}{\mathcal{D}^{ER}}
\section{Continual Learning under Experience Replay (ER)}

\begin{figure}[t]
  \centering
\includegraphics[width=0.99\textwidth]{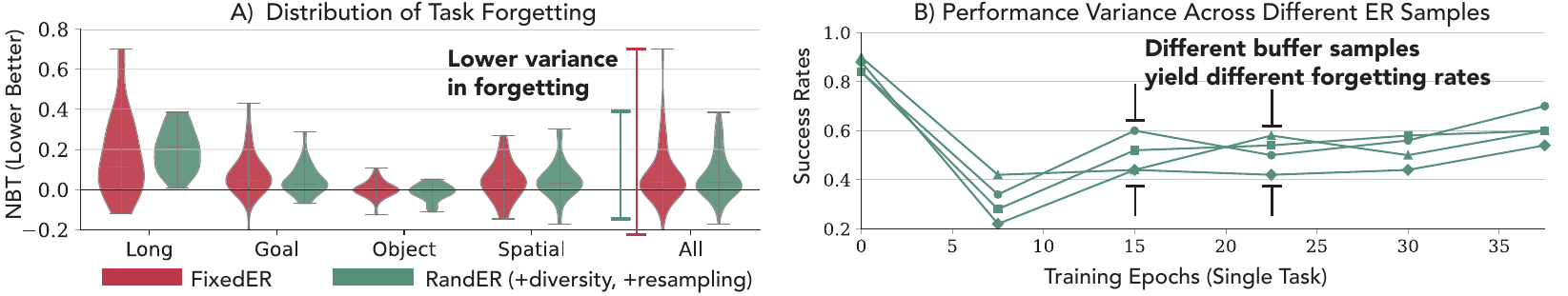}
 \caption{ \textbf{Sensitivity of Experience Replay (ER) to Data Selection.} The standard ER practice (\naivemethod{}) samples at the trajectory level and does not resample $\der$ after a new task. If we sample uniformly at the \textit{transition} level and resample $\der$ for every new task (\basemethod{}), we get more consistent performance across tasks (A). However, remaining variance exists between samples of $\der$ (B), indicating sensitivity to the exact data present in $\der$.}
  \label{fig:exp1}
    \vspace{-5mm}
\end{figure}  

We consider the continual learning (CL) setting where a series of tasks $\{T^1, T^2,...,T^N\}$ and their associated datasets $\{\mathcal{D}^1, \mathcal{D}^2, ..., \mathcal{D}^N\}$ are given to a robot policy in sequence. The continual learning objective is to keep acquiring new tasks without forgetting past ones \cite{lesort2020continual}. The magnitude of forgetting is measured through \uline{Negative Backward Transfer (NBT)}, a measure of \textit{stability} computed as the success drop of task $T^k$ as the policy learns $T^{k+1}...T^N$ \cite{liu_libero_2023}. NBT is a direct measurement of forgetting.
%
To reduce NBT, the ER method~\cite{chaudhry_tiny_2019} approaches this catastrophic forgetting problem by keeping an additional buffer $\der$ of past task data, usually a fixed size or a small proportion of the past task data. While training task $T^k$, data is sampled from both $\der$ and the current task set $\mathcal{D}^k$. In the following experiments, we investigate properties of ER in continual robot learning that motivate Memory Anchors. For other metrics used in continual learning, refer to Appendix \ref{app:metrics}.

\subsection{Study: Different Sampling Approaches for ER Buffer}
\label{sec:rand_exp}
\begin{questionbox}
\textbf{Question:}
Is continual learning performance sensitive to the choice of data in the ER buffer?
\end{questionbox}

Many sampling approaches exist for ER in continual learning applications, both for creating $\der$ and sampling from $\der$ \cite{prabhu2023computationallybudgetedcontinuallearning, chaudhry_tiny_2019}. In the robotic setting, prior works have populated $\der$ in different ways, including episode sampling \cite{liu_libero_2023, zhu_can_2026} and transition sampling \cite{liu2026pretrained}, but only simple statistics like buffer size and sampling frequency have been compared \cite{liu_libero_2023, liu2026pretrained,zhu_can_2026}. 
To motivate the presence of individually important data in $\der$, we compare some of these design choices and look at how $\der$ \textit{diversity} impacts continual learning performance. We consider two random $\der$ selection strategies: \naivemethod{} \cite{liu_libero_2023}, which contains correlated episode-level samples, and \basemethod{}, which adds diversity by sampling at the \textit{transition} level and resampling $\der$ after every task. 

\textbf{Experiment Suites.} For our simulation experiments below and in later sections, we evaluate on the LIBERO setup, a set of four sequential task suites (LIBERO-Long, LIBERO-Goal, LIBERO-Object, LIBERO-Spatial) on a tabletop manipulation environment, each consisting of ten tasks trained in sequence \cite{liu_libero_2023}. 
We train a diffusion policy \cite{chi_diffusion_2023} through behavior cloning on the provided datasets, using ER regularization \cite{chaudhry_tiny_2019}. For more details, refer to Appendix \ref{app:details}. 

\textbf{Diverse $\der$ reduces variance in task forgetting.} We plot the forgetting (NBT) of each task across three training permutations (30 total NBT values per suite) in \cref{fig:exp1}A. The average NBT values of \basemethod{} and \naivemethod{} are comparable, but \basemethod{} exhibits $36$\% lower variance across tasks and training permutations. This lower variance also gives \basemethod{} better worst-case performance, with 25\% lower worst-case forgetting compared to \naivemethod{}. 

\textbf{Past task retention varies by $\der$ sample.} The reduced variance and worst-case forgetting of \basemethod{} could be explained by sensitivities of some tasks to $\der$ quality. We take a closer look at data sensitivity by isolating one task $T^j$ that encounters significant catastrophic forgetting after training on task $T^k$. Starting from the same checkpoint, we train $T^k$ multiple times with \basemethod{}, varying only the sample of $\der$ (\cref{fig:exp1}B). With task ordering constant, the level of forgetting still varies up to $20$\% between $\der$ samples.


These two results provide evidence that robot continual learning is sensitive to the types of memories sampled in $\der$. Appearing through random selection in \basemethod{}, these memories appear to have a strong (20\%) impact on task retention. Finding these memories requires understanding the mechanism behind forgetting, which we explore in the next section.


\begin{keytakeaway}
\textbf{Takeaway:} Continual learning performance is sensitive to the selected ER buffer, especially for the worst-case tasks for catastrophic forgetting.
\end{keytakeaway}

\subsection{Task Relationships in Catastrophic Forgetting}
\label{sec:taskrelationships}

\begin{figure}[t]
  \centering
  \includegraphics[width=0.99\textwidth]{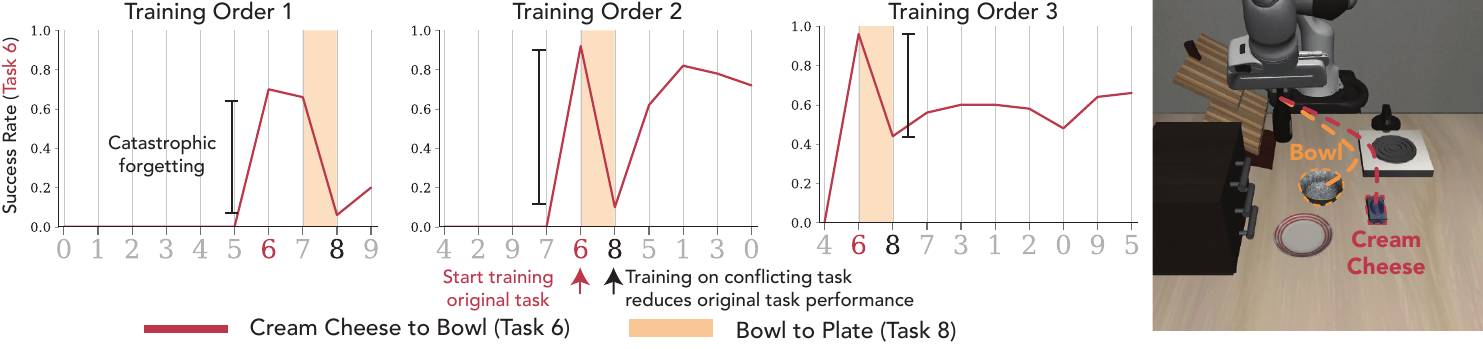}
 \caption{ \textbf{Task Conflicts Persist Across Permutations.} With an ER regularization (\basemethod), task performance stays steady for some new tasks and drops significantly for others. Shown above, \texttt{Cream Cheese to Bowl} drops in performance when the policy learns an additional task involving picking the Bowl instead.}
  \label{fig:taskint}
    \vspace{-5mm}
\end{figure}  

\begin{questionbox}
\textbf{Question:}
What makes a trained task encounter significant catastrophic forgetting? 
\end{questionbox}
Using an ER buffer normally enables most tasks to maintain their performance, degrade slightly, or even improve over time (\cref{fig:exp1}A). However, there are some tasks like in \cref{fig:exp1}B (\texttt{Cream Cheese to Bowl}), which drop performance significantly during new task training. In this section, we consider why this might happen. 
Instead of sampling different $\der$ for one new task, we look at performance across a ten-task training run for three different training orders.

\textbf{Certain task-task interactions contribute to the majority of forgetting.} Plotting the performance of \texttt{Cream Cheese to Bowl}, we see that forgetting happens abruptly at task borders, particularly when \texttt{Bowl to Plate} is learned, an impact persisting across training orders (\cref{fig:taskint}). This interaction pair yields an average performance loss of $24$\%, amounting to more than $5$x the average NBT for the LIBERO-Goal task suite. This example reflects a general trend: across three training orders of LIBERO-Goal, \textit{the total forgetting of only 2 tasks (out of 10) accounted for more than 50\% of all negative backward transfer.} Furthermore, across the 80 task interaction pairs present in the training orders for LIBERO-Goal, the worst 5 task interaction pairs caused an average of 20\% success drop, while the other 75 pairs caused an average 0.1\% success \textit{rise}, indicating the disproportionate contributions of certain task-task conflicts to overall forgetting. For more analyses, see Appendix \ref{app:metrics_tasks}.

\begin{wrapfigure}{r}{0.4\textwidth}
\vspace{-6mm}
\centering
\includegraphics[width=0.99\linewidth]{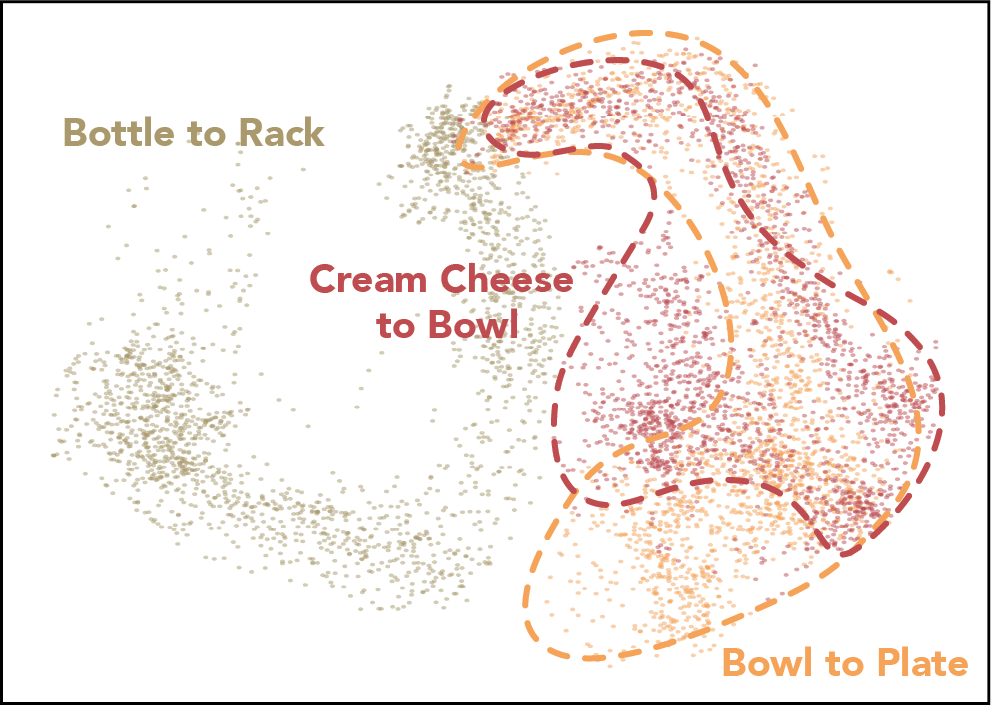}
\vspace{-4mm}
\caption{\textbf{New Data Overlaps Past Tasks.} When new and old tasks share visual or semantic similarities, a policy can collapse a new task (\textcolor{mypastelred}{red}) into an existing task's representation space (\textcolor{mypastelorange}{orange}), causing conflict.}
\label{fig:repressentations}
\vspace{-6mm}
\end{wrapfigure}

\textbf{Conflicting task pairs share observation representation overlap.}  Qualitatively, when the policy forgets \texttt{Cream Cheese to Bowl} upon training \texttt{Bowl to Plate}, it starts to grab the bowl instead of the cream cheese (\cref{fig:taskint}). We hypothesize that these two tasks overlap in the visual-language embedding space due to their physical closeness and the presence of ``\texttt{bowl}'' in both task instructions, leading to confusion over the actions for each task. Indeed, we see this in a PCA analysis (\cref{fig:repressentations}).

Although the two tasks share similar trajectories while reaching towards the objects, they also have opposing action labels in other subtrajectories. This combination, \textit{\textbf{overlapping representations (input) and different actions (output)}}, can lead to interference in the shared model resources and cause catastrophic forgetting. Old task data sampled in this overlap during new task training can counteract this effect by acting as a regularizer. We hypothesize that the chance appearance of this special subset through \basemethod{}'s random selection is a driving factor in the performance variations seen in \cref{fig:exp1}B. We identify this data as our \textit{Memory Anchors}. To test our hypothesis, we now propose a method of finding these anchors during the training process.


\begin{keytakeaway}
\textbf{Takeaway:} A new task likely interferes with an old task if they share \uline{similar observation spaces} but require \uline{different action strategies}.
\end{keytakeaway}

\section{Memory Anchors}

\label{sec:findingmemoryanchors}

Using our previous findings, we propose a three-step method for finding Memory Anchors $\mathcal{D}^{Anchor}$ from past task data $\{\mathcal{D}^1, ..., \mathcal{D}^{n-1}\}$ using the new dataset $\mathcal{D}^n$ and the current policy $\pi_\theta^{n-1}$ trained on past task data. 

\textbf{Step 1: Find State Representation Overlap.} Before new task training, the existing policy may already represent new task data in the same latent region as a conflicting old task, collapsing the features that distinguish them, 
as demonstrated in \cref{fig:repressentations}. 
To find this subset of new data $\mathcal{D}^n$, we place $\mathcal{D}^n$ into the observation latent space of the policy $\pi_\theta^{n-1}$. Then, we use the latent distance of $\mathcal{D}^n$ to $\{\mathcal{D}^1, ..., \mathcal{D}^{n-1}\}$ to locate all $s \in \mathcal{D}^n$ within the manifold of past task data. We populate a set $\mathcal{D}^n_{S_{ID}}$ with these data points.

\begin{figure}[t]
  \centering
  \includegraphics[width=0.95\textwidth]{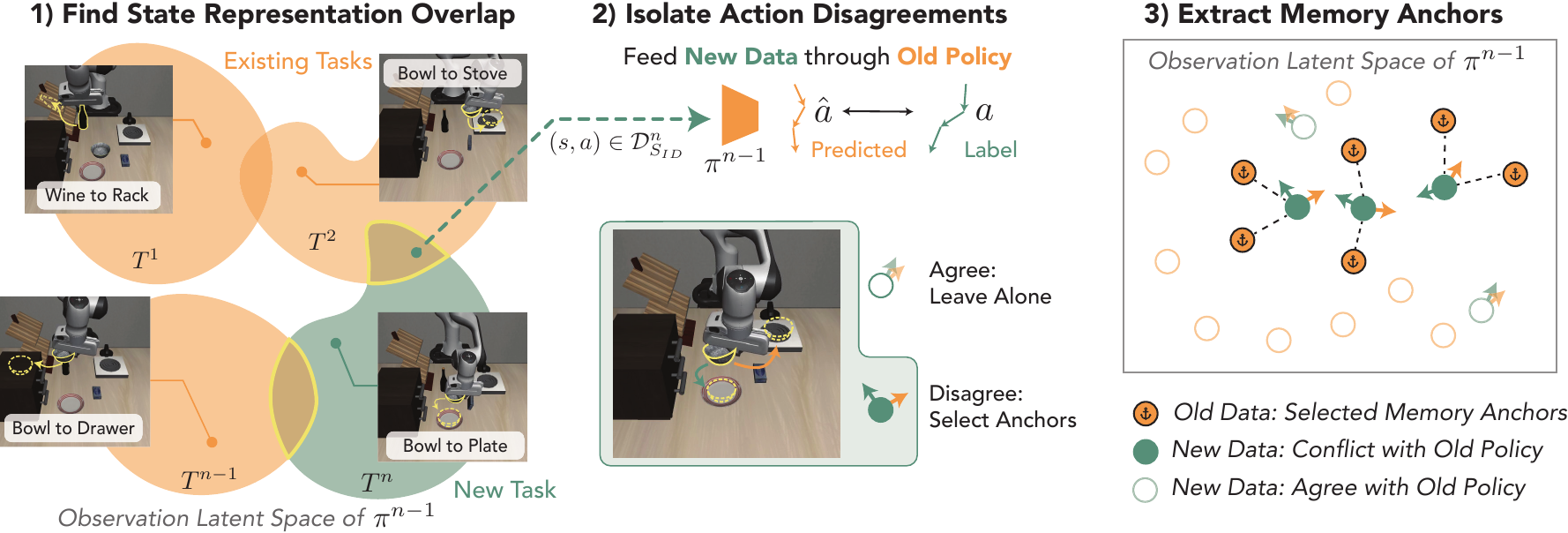}
  \caption{\textbf{Finding Memory Anchors}. From the overlapping latent state spaces of old and new task policy~(1), we isolate new task points yielding the highest action disagreement with the old data (2). To do this, 
  we predict new task actions using the current policy (2 Upper) and compare to the action labels to measure disagreement (2 Lower). Using these points, we extract old task data that exhibits the most similarity (3). These Memory Anchor candidates have state representation similarities and action disagreement with the new task data (\S \ref{sec:findingmemoryanchors}).
  }
  \label{fig:method}
    \vspace{-5mm}
\end{figure}

\textbf{Step 2: Isolate Action Disagreements through Generative Policy.} From our state-similar data $\mathcal{D}^n_{S_{ID}}$, we want to find a smaller subset $\mathcal{D}^n_{OD} \subset \mathcal{D}^n_{S_{ID}}$ that requires different action outputs than the old data. The old data will not contain an exact match for states in $\mathcal{D}^n_{S_{ID}}$, making retrieval-based action comparisons difficult. Instead, we propose querying the current policy $\pi_\theta^{n-1}$ directly. 
From $(s, a) \in \mathcal{D}_{S_{ID}}^n$, we first add gaussian noise to $a$ to make $\tilde{a}$. Then, we denoise with the policy $\hat{a}\sim \pi_\theta^{n-1}(\cdot | s, \tilde{a})$. This noise-denoise approach prevents false positives with action multimodality.
If $\|\hat{a} - a\| \gg 0$, then this $(s, a)$ sample of new data will require a significant change in the policy during training, likely causing conflict with existing knowledge. 
We populate $\mathcal{D}^n_{OD}$ with these action-disagreeing data points (\cref{fig:method} Middle). For more implementation details, refer to Appendix \ref{app:details_anchors} 


\textbf{Step 3: Extract Memory Anchors from Old Data.} As hypothesized (\S \ref{sec:taskrelationships}), the old data most similar to $\mathcal{D}^n_{OD}$ serves an important regularization role during new task training with ER. Thus, we find the data from $\{\mathcal{D}^1, ..., \mathcal{D}^{n-1}\}$ that is most similar to $\mathcal{D}^n_{OD}$ in the representation space of $\pi_\theta^{n-1}$ (\cref{fig:method} Right). This extracted data is $\mathcal{D}^{Anchor}$: the Memory Anchors. For qualitative examples of Memory Anchors, refer to Appendix \ref{app:realrobot_anchors}. For a visual overview of this approach, refer to \cref{fig:method}.

\section{Impact of Memory Anchors on Forgetting}
\label{sec:experiments}



\subsection{Reducing Access to Memory Anchors}
\label{sec:removeanchors}
\begin{figure}[t]
  \centering
  \includegraphics[width=0.95\textwidth]{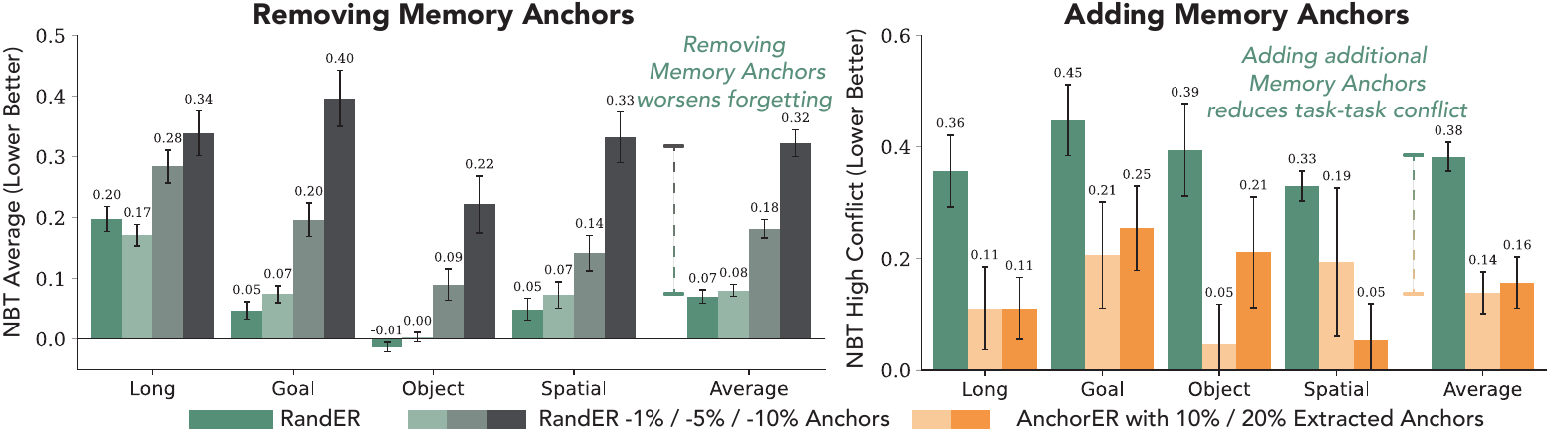}
  \caption{\textbf{Impact of Memory Anchors on Forgetting}. Reducing access to a small proportion of the best Memory Anchors (\S \ref{sec:removeanchors}) during the ER sampling process leads to large increases in catastrophic forgetting~(Left). Symmetrically, increasing the concentration of Memory Anchors in an ER buffer reduces forgetting on high-conflict task pairs (Right). 
  }
  \label{fig:mainresults}
    \vspace{-5mm}
\end{figure}

We hypothesized that Memory Anchors play a key role in maintaining past task performance. Therefore, reducing the concentration of these anchors should increase forgetting. In the following experiment (\cref{fig:mainresults} Left), we remove $1$\%, $5$\%, and $10$\% best memory anchors (lowest latent distance to $\mathcal{D}^n_{OD}$) from past task data and sample an ER buffer of $1000$ memories from the reduced dataset. 


With a randomly selected ER buffer, all task suites except for LIBERO-Long were able to get a low NBT with the $1000$ memory budget. However, without the ability to sample the top 10\% of Memory Anchors, \textit{forgetting increased more than 4.5x}. Most surprisingly, forgetting still increased in some cases when only 1\% of the past task dataset was excluded. These results show that our Memory Anchor candidates are behaving like we hypothesized: a small subset of data plays a critical role in maintaining past performance.  For more analysis, refer to Appendix \ref{app:metrics_memanchors}.

\begin{keytakeaway}
\textbf{Takeaway:} Removing even a small proportion of Memory Anchors from the ER selection process causes large increases in task forgetting.   
\end{keytakeaway}

\subsection{Adding More Memory Anchors}
\label{sec:addanchors}
Memory Anchors naturally exist in ER buffers, and the experiment in the last section demonstrated their large role in task retention. But will \textit{adding more} Memory Anchors boost retention? We test this idea through \methodname{}: an ER approach that fills $n$\% of the buffer with Memory Anchors extracted using our method. We conducted experiments on small ER buffers (1\% of past task datasets) to expose more catastrophic forgetting and task-task conflicts, 
which we measure by isolating the six most conflicting task pairs for each suite and computing their forgetting on \basemethod{} and \methodname{}. By adding more Memory Anchors, \textit{the average forgetting on these high-conflict tasks was reduced by more than $63$\%} (\cref{fig:mainresults} Right).



Although all suites are sensitive to Memory Anchors (\S \ref{sec:removeanchors}) and adding more Memory Anchors improves performance on high task conflicts, \methodname{} also provides a measurable ($>1$ SEM) boost in overall task retention on LIBERO-Goal by reducing the \uline{average} NBT by 37\% (Appendix \ref{app:metrics_anchorer}). Unlike the other suites that have different environments and/or objects between tasks, Goal is \textit{homogeneous} with the same scenes and objects. It reflects a more realistic continual learning setup with many skills learned in the same environment with similar objects. Homogeneous tasks also have the most observational overlap and conflict, making them difficult to learn and more affected by Memory Anchor presence. 

\begin{keytakeaway}
\textbf{Takeaway:} Including more Memory Anchors mitigates high-conflict task interactions. For task sequences with similar objects and environments, it also improves overall task retention. 
\end{keytakeaway}

\subsection{VLA Sensitivity to Memory Anchors}
\label{sec:vla}

\begin{wraptable}{r}{0.5\textwidth}
\vspace{-3.5em}
\centering
\caption{\textbf{Impact of Memory Anchors on VLA. } The availability of Memory Anchors influences the continual learning performance of the $\pi_{0.5}$ VLA.}
\label{tab:vla_results}
    \small 

\begin{subtable}[t]{\linewidth}
    \centering
    \caption{Reducing Memory Anchors (NBT Average).}
    \begin{tabular}{lcc}
        \toprule
        RandER & RandER -$5\%$ & RandER -$10\%$ \\
        \midrule
        \textbf{0.12 $\pm$ 0.02}  & 0.20 $\pm$ 0.04 & 0.35 $\pm$ 0.04  \\
        \bottomrule
    \end{tabular}
\end{subtable}

\vspace{0.6em}

\begin{subtable}[t]{\linewidth}
    \centering
    \caption{Adding More Memory Anchors (NBT Average).}
    \begin{tabular}{lcc}
        \toprule
        RandER & AnchorER 10\% & AnchorER 20\%  \\
        \midrule
        0.45 $\pm$ 0.05  & \textbf{0.35 $\pm$ 0.04} & \textbf{0.36 $\pm$ 0.04}  \\
        \bottomrule
    \end{tabular}
\end{subtable}
\vspace{-1.5em}
\end{wraptable}

The results of the previous experiments demonstrate the influence of Memory Anchors on diffusion policies trained from scratch. Past works have discovered that Vision Language Action (VLA) models have better continual learning abilities, reaching near-zero forgetting with a large ER buffer size (Appendix \ref{app:vla}) \cite{liu2026pretrained}. We extend these investigations by looking at VLA sensitivity to Memory Anchors on our experiment setup, where the ER buffer is small relative to the past dataset size. Proportionally-small buffers represent a practical setup with large pretraining sets or long task sequences. We conduct the same subtraction (\S \ref{sec:removeanchors}) and addition (\S \ref{sec:addanchors}) experiments on the $\pi_{0.5}$ VLA in LIBERO-Goal. The results are shown in Table \ref{tab:vla_results}.  

On the subtraction experiment with $1000$ total memories, we find that $\pi_{0.5}$ exhibits \textit{worse} forgetting than from-scratch diffusion policy on a randomly-selected ER buffer (0.12 compared to 0.05). Taking away Memory Anchor access further increased forgetting by 2.9x, consistent with results in \S \ref{sec:removeanchors}. On the addition experiment with the 1\% memory buffer, we also find worse forgetting than from-scratch diffusion (0.45 compared to 0.18). Enriching the buffer with Memory Anchors reduces forgetting by 22\%, which is consistent with the results in \S \ref{sec:addanchors}. Experimental details are in \ref{app:vla_details}.

\begin{keytakeaway}
\textbf{Takeaway:} On proportionally smaller ER buffers, VLAs still exhibit catastrophic forgetting. Under these conditions, Memory Anchors impact their continual learning performance.
\end{keytakeaway}


\begin{figure}[t]
  \centering
  \includegraphics[width=0.99\textwidth]{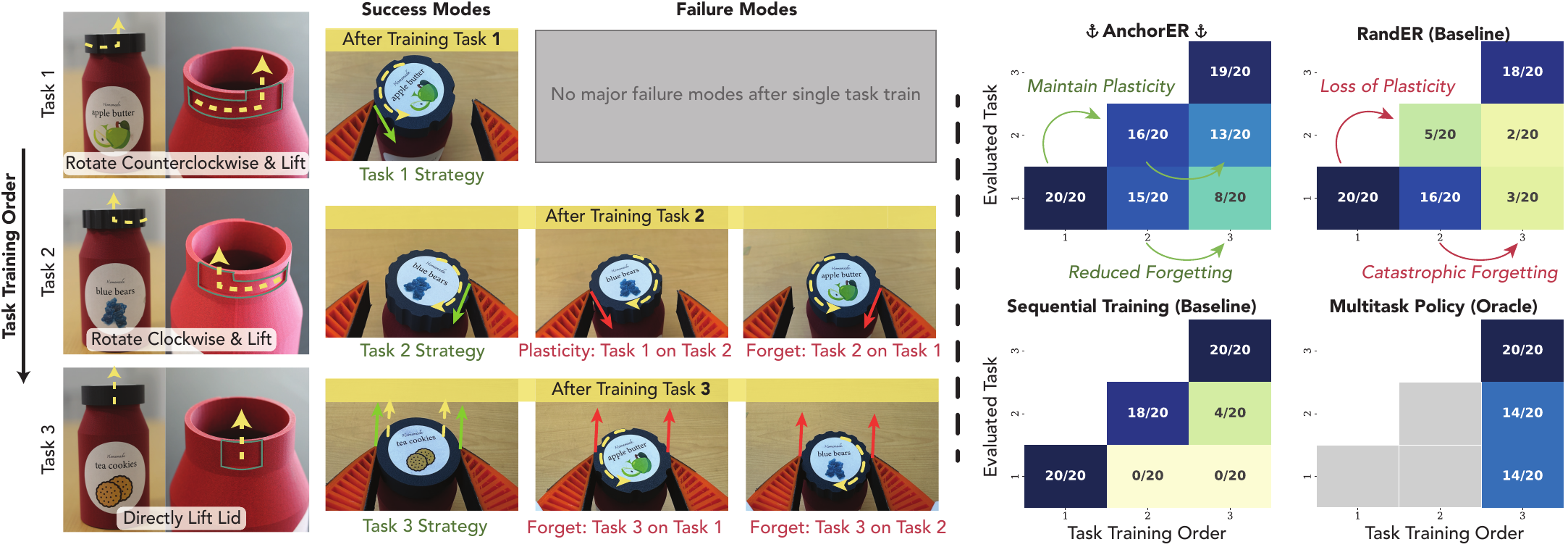}
\caption{\textbf{OpenJar Real Robot Task Suite}. We propose a set of three tasks with visually similar jars that require opposing strategies to open (Left), leading to many instances of catastrophic forgetting and plasticity loss when trained improperly (Middle). Using \methodname{}, the robot achieves less forgetting and 1.7x higher overall success rate compared to a randomly-sampled ER buffer, \basemethod{} (Right).}
  \label{fig:realrobot}
    \vspace{-5mm}
\end{figure}

\subsection{Memory Anchors on a Real Robot}
\label{sec:realrobot}


The \textit{homogeneous} tasks in LIBERO-Goal represent a realistic setup: multiple tasks involving similar objects in the same deployment environment. We bring these challenges to a real robot through the \textbf{OpenJar} task suite (\cref{fig:realrobot}). \textbf{OpenJar} consists of perceptually similar jars that require very different strategies to open (counterclockwise, clockwise, direct lift). Because the robot might touch the jars multiple times to unscrew their lids, these tasks require many decision points per trajectory (\cref{fig:realrobot}). Their distinguishing visual features are separate from the jar opening affordances, making the jars likely to be initially overlapping in their representations. We use the Universal Manipulation Interface (UMI) \cite{chi2024universalmanipulationinterfaceinthewild, patel2026behaviorpromptingpolicydemonstrations} for data collection and deployment on an ARX robot arm \cite{gao2026gatedmemorypolicy}.

On a sequential baseline, the robot overgeneralizes on the current jar and exhibits nearly complete forgetting (\cref{fig:realrobot} Right). Adding uniform ER (\basemethod{}) reduces forgetting, but introduces sensitivity to the buffer size. At a buffer size large enough to preserve the first task, the policy loses plasticity for the second task (\cref{fig:realrobot} Right). Applying \methodname{} not only prevents forgetting of the second task while learning the third, but it \textit{also enables the second task to learn properly} by focusing on the critical decision points within the trajectory.

\begin{figure}[t]
  \centering
  \includegraphics[width=0.99\textwidth]{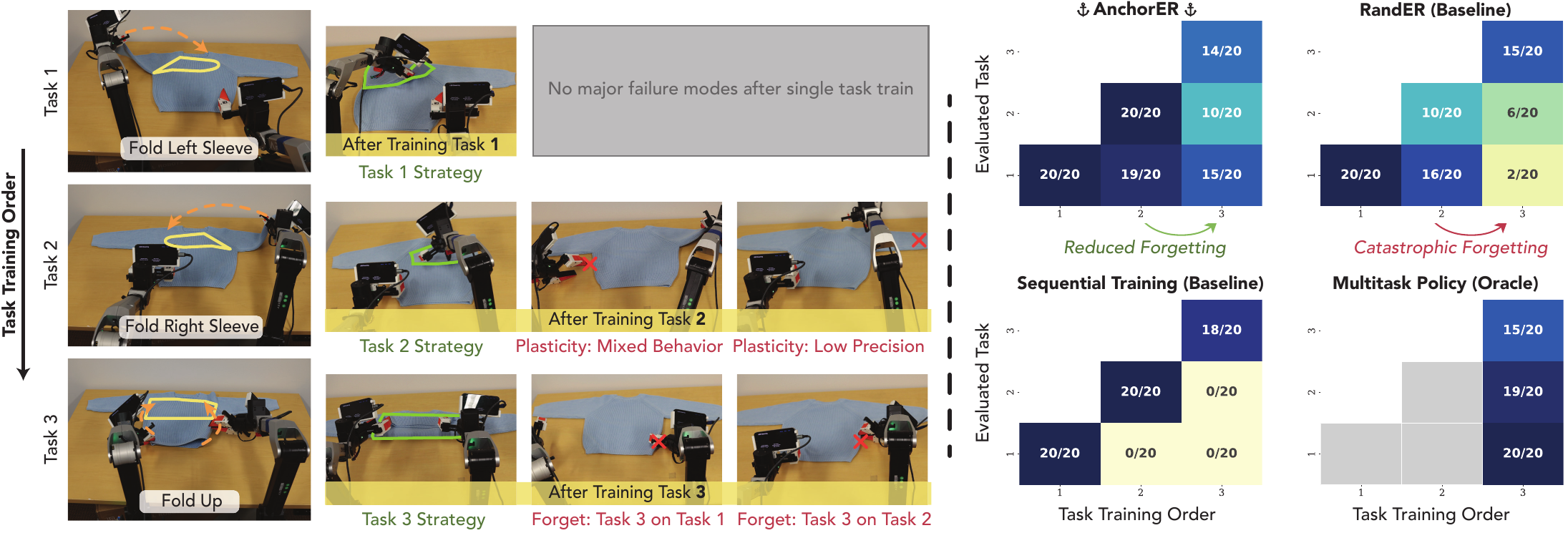}
\caption{\textbf{SweaterFold Real Robot Task Suite}. We further test continual learning strategies on a realistic set of language-conditioned garment-folding tasks. Consistent with the results in \textbf{OpenJar}, we find that \methodname{} achieves a better balance of plasticity and task retention.}
  \label{fig:sweater}
    \vspace{-5mm}
\end{figure}

These trends continue in a second task suite, \textbf{SweaterFold} (\cref{fig:sweater}). This suite consists of three bimanual, language-conditioned sweater folding behaviors (fold left, fold right, fold up) learned in sequence. It represents a realistic example of continual learning in the real world. Like before, adding uniform ER improves past task retention, but catastrophic forgetting still happens after the third task, and the second task is challenging to learn. In comparison, \methodname{} is able to balance plasticity and task retention, achieving 1.7x higher final success rates than \basemethod{} on the same ER budget. For details and qualitative visuals of the Memory Anchors, refer to Appendix \ref{app:realrobot_details}. 







\begin{keytakeaway}
\textbf{Takeaway:} On two suites of highly conflicting real-robot tasks, adding Memory Anchors (\methodname{}) improves both backward and forward transfer. 
\end{keytakeaway}

\subsection{Baselines and Ablations}
We have presented \methodname{} as an improved ER selection method and shown its benefits on simulated and real environments. To quantify \methodname{} against other selection and buffer-based training methods, we compare with two baselines: \textbf{Maximal Interference Retrieval} (MIR) \cite{aljundi2019online} and \textbf{Gradient-Based Sample Selection} (GSS) \cite{aljundi2019gradient}. We also add an ablation: \textbf{removing the action disagreement (Step 2)} from the AnchorER pipeline to test a simpler, nearest-neighbor retrieval process. We present average NBT results on LIBERO-Goal in Table \ref{tab:baselines}.

\begin{wraptable}{r}{0.5\textwidth}
    \vspace{-1.5em}
 \small
    \caption{\textbf{Baselines.} \methodname{} selection outperforms relevant baselines and ablations. A correlation also exists between the number of Memory Anchors in a method's buffer and its task retention ability.}
    \centering
    \label{tab:method_comparison}
    \setlength{\tabcolsep}{3pt}
    \resizebox{0.5\textwidth}{!}{%
    \begin{tabular}{lccccc}
        \toprule
       & \textbf{Ours} & \textbf{MIR} & \textbf{GSS} & \textbf{Ours-ActionDis} & \textbf{RandER} \\
        \midrule
        Avg. NBT $\downarrow$ & \textbf{.11$\pm$.01} & .14$\pm$.01 & .17$\pm$.03 & .13$\pm$.01 & .18$\pm$.02 \\
        Anchor Conc. $\uparrow$ & .27 & .13 & .08  & .25  & .10 \\
        \bottomrule
    \end{tabular}%
    }
    \label{tab:baselines}
    \vspace{-1.2em}
\end{wraptable}

The Memory Anchor enrichment method (\methodname{}) still yields the lowest average NBT. GSS only relies on gradient diversity to curate the buffer, not the relationship to the new task like \methodname{}. MIR does consider this relationship, but it only influences the \textit{sampling} from a random buffer instead of the curation process. The ablation demonstrates that action disagreement does contribute to \methodname{}. Without it, nearest-neighbor selection also pulls in shared subtrajectories (e.g. reaching for the bowl in GOAL), which are already rehearsed by the new data. 

To unify these findings under our Memory Anchor hypothesis, we measured a \textit{Memory Anchor Concentration} as the proportion of top (10\%) Memory Anchors found in the most sampled fraction (40\%) of each buffer. Policy performance and Memory Anchor concentration are positively correlated, further supporting our claim that Memory Anchors are driving past task retention.

\begin{keytakeaway}
\textbf{Takeaway:} \methodname{} outperforms other buffer enrichment methods, and even across these other methods, the presence of Memory Anchors is still correlated with task retention.
\end{keytakeaway}

\label{sec:baselines}





	

\section{Conclusion and Discussion}
\label{sec:conclusion}
In this paper, we introduced the idea of \textit{Memory Anchors}, a small subset of past task data critical for maintaining performance during sequential training with Experience Replay (ER). We proposed a method for identifying Memory Anchors and enriching the ER buffer with them. Through a series of studies in simulation and on a real robot, we demonstrated the importance of Memory Anchors in continual learning for overall task performance, high-conflict tasks, and small buffer sizes. 

\textbf{Limitations:}
Our Memory Anchor extraction process requires access to the current policy's latent space and its past training data, which may not be available for proprietary models. It also requires computing representations of these past datasets at the start of every task, which is computationally intensive for larger sets. It is possible to use alternative latent spaces and open-source data outside of the policy's past training set for Memory Anchors, which we leave to future work. We also test continual learning on a relatively short sequence of 3-10 tasks, which may not represent the nature of continuous policy deployment in the real world. 


\clearpage

    \acknowledgments{Maximilian Du is supported by the Knight-Hennessy Fellowship and the NSF Graduate Research Fellowship Program (GRFP). This work was supported in part by Toyota Research Institute, NSF Award \#2143601, \#2037101, and \#2132519. We would like to thank Austin Patel and Yihuai Gao for assisting with the policy deployment on the ARX arm. We appreciate all members of the REAL lab at Stanford for their detailed feedback on paper drafts and experiment directions. The views and conclusions contained herein are those of the authors and should not be interpreted as necessarily representing the official policies, either expressed or implied, of the sponsors.}



\bibliography{ref}  

\newpage 
\appendix

\section{Real Robot Experiments: Additional Results and Analysis}


\label{app:realrobot}

\subsection{Real Robot Details: OpenJar}
\label{app:realrobot_details}
We consider a set of three identically-shaped jars that take distinctive strategies to open. The first jar lid (Task 1, \textbf{Counterclockwise}) requires a 90 degree counterclockwise rotation before the lid can be lifted. The second jar lid (Task 2, \textbf{Clockwise}) requires a 90 degree clockwise rotation before the lid can be lifted. The third jar lid (Task 3, \textbf{Lift}) does not allow any rotation and requires a direct lift to remove. The jars are distinguished by their label text and icon, as well as the texture present on the lid. At least one of these features are visible from the eye-in-hand camera at all times, allowing the task to be determined purely from visual observation without history or task conditioning.

\begin{wrapfigure}{r}{0.35\textwidth}
\vspace{-5mm}
\centering
\includegraphics[width=0.99\linewidth]{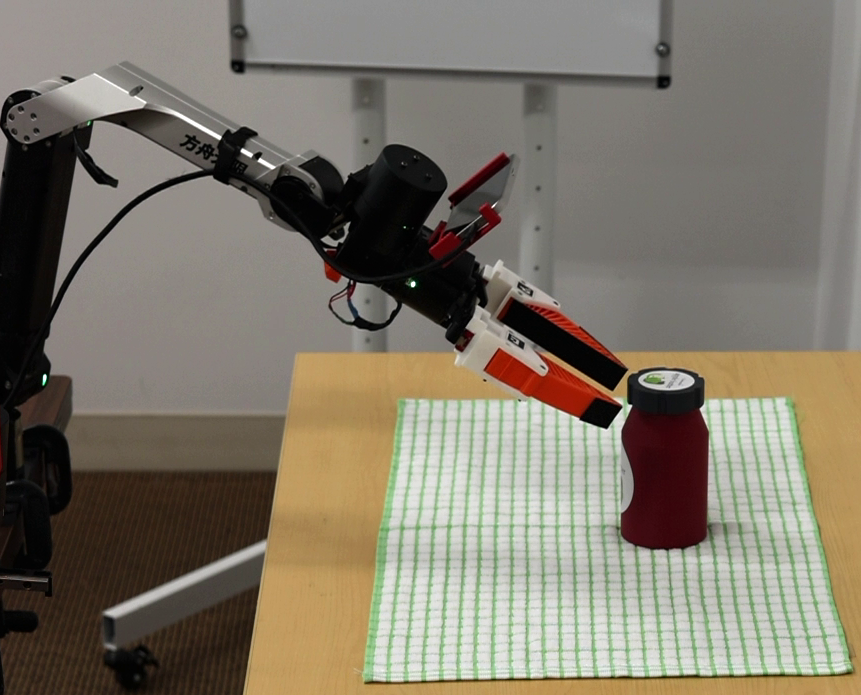}
\vspace{-2mm}
\caption{\textbf{Robot Setup.} We use the UMI setup to collect data and then deploy a trained policy on an ARX robot arm for \textbf{OpenJar} tasks.}
\label{fig:robotsetup}
\vspace{-4mm}
\end{wrapfigure}

We use the UMI setup on an ARX arm \cite{chi2024universalmanipulationinterfaceinthewild, gao2026gatedmemorypolicy}. We collect demos using an UMI-style data collection device with an iPhone \cite{patel2026behaviorpromptingpolicydemonstrations} as a camera. We start the gripper above and behind the jar, moving down and forward until the gripper is aligned with the jar lid (neutral position). For \textbf{Counterclockwise}, we move the \textit{left} gripper until it makes contact with the left edge of the jar and we move the gripper back, causing the jar to rotate. We pull back to the neutral position and repeat until the jar is sufficiently rotated. Then, we move forward and grasp the lid, lifting it. The \textbf{Clockwise} task employs the same strategy but using the \textit{right} gripper contact to rotate the lid clockwise. Finally, the \textbf{Lift} task goes directly to lid grasping and lifting without a rotation. We designed this demo behavior with task homogeneity in mind: the overall observations between the tasks are very similar but the requested actions are different. We collect 124 demos for \textbf{Counterclockwise}, 120 demos for \textbf{Clockwise}, and 148 for \textbf{Lift}. 

We train a u-net diffusion policy \cite{chi_diffusion_2023}, using a TIMM VIT encoder (\verb|vit_base_patch16_clip_224|) on narrow-view RGB and wide-view RGB camera streams. For both \methodname{} and \basemethod{}, we used an ER buffer size of 5000 transitions. We trained the first two tasks for $100$ epochs and the last task for $40$ epochs due to its simplicity compared to the first two. Across all conditions, we chose the last epoch for each task in our evaluations. For evaluation, we randomized the position of the jars in the workspace. 

\subsection{Memory Anchors Visualized for \textbf{OpenJar}}
\label{app:realrobot_anchors}
\begin{figure}[t]
  \centering
  \includegraphics[width=0.95\textwidth]{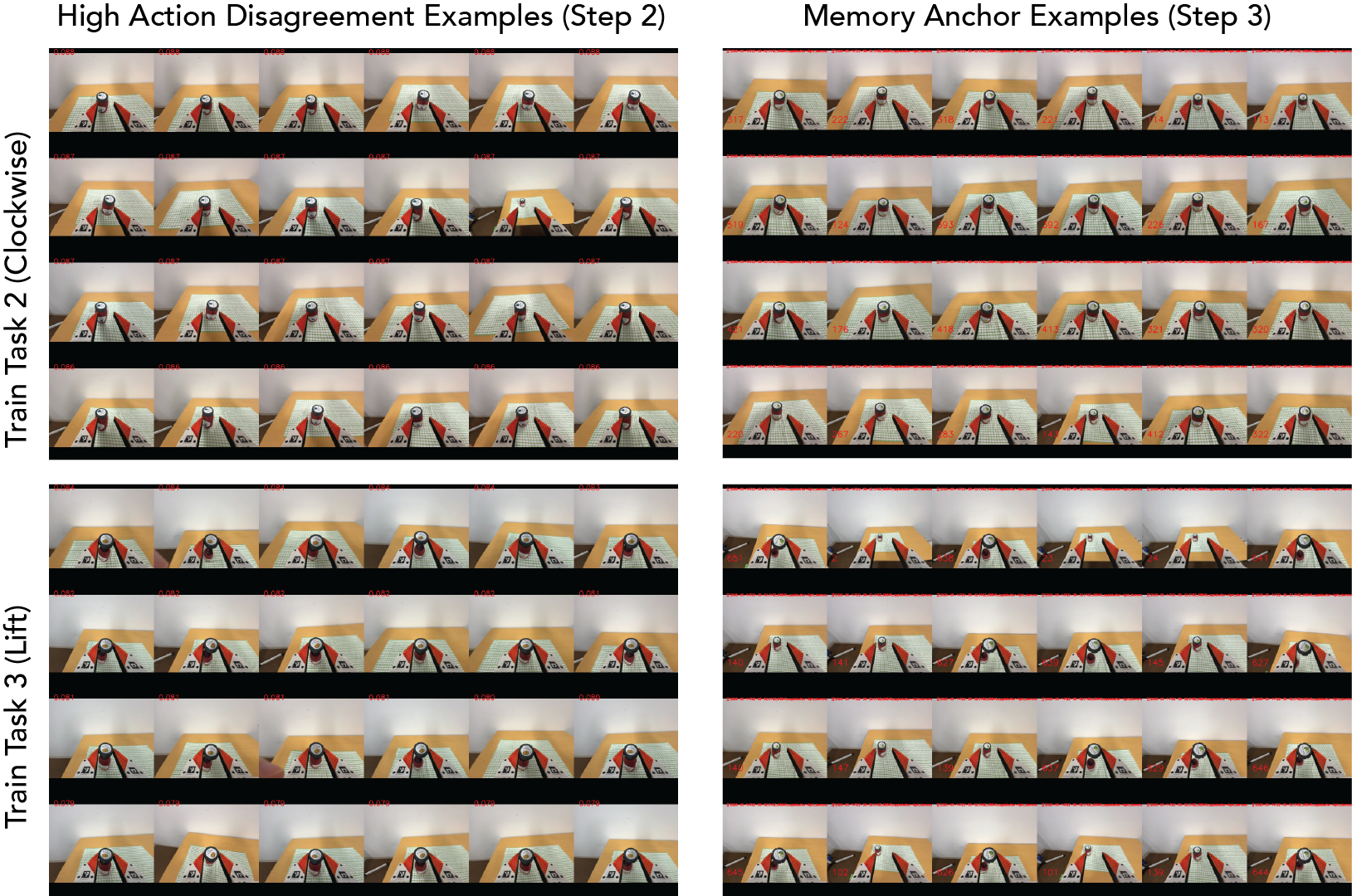}
  \caption{\textbf{Examples of Action Disagreement and Selected Memory Anchors in Real-World Task}. High action disagreement in the \textbf{OpenJar} task happens near the jar lid contact (Left column), and the corresponding Memory Anchors also focus on the jar lid contact of the old tasks (Right column). The images are sampled from the top $100$ action disagreement and memory anchors, respectively. 
  }
  \label{fig:app_anchorsvisual}
    \vspace{-5mm}
\end{figure}

The memory anchor extraction process for \textbf{OpenJar} is intuitive. As seen in \cref{fig:app_anchorsvisual}, the points in the new data with high action disagreement (Step 2, \S \ref{sec:findingmemoryanchors}) are near the jar lid contact. The resulting Memory Anchors are also focused around this contact region (\cref{fig:app_anchorsvisual} Right).

When the robot learns Task 2 (\textbf{Clockwise}, \cref{fig:app_anchorsvisual} Upper Row), the action disagreement (\cref{fig:app_anchorsvisual} Upper Left) happens on the final approach to the lid. This is because the old policy on Task 1 (\textbf{Counterclockwise}) would move right to contact the left gripper on the jar lid for the counterclockwise rotation. However, the new data on Task 2 (\textbf{Clockwise}) requires the robot to move \textit{left} to contact the right gripper on the jar lid. When the robot gripper is near the center, the visual representations are overlapped between \textbf{Clockwise} and \textbf{Counterclockwise} but the actions are different, leading to action disagreement. The retrieved data (\cref{fig:app_anchorsvisual} Upper Right) focuses on this exact region on \textbf{Counterclockwise}. The presence of this matching old task data in the region of action disagreement makes \methodname{} more successful on this task suite.

The same effect can be seen when the robot learns Task 3 (\textbf{Lift}, \cref{fig:app_anchorsvisual} Lower Row). Here, the action disagreements happen slightly later, when the gripper closes and the robot lifts (\cref{fig:app_anchorsvisual} Lower Left), and the retrieved Memory Anchors also focus on similar lifting behavior as well as some early reaching behavior (\cref{fig:app_anchorsvisual} Lower Right). 

\subsection{Additional Qualitative Results: OpenJar}
\label{app:realrobot_qualitative}

\begin{wrapfigure}{r}{0.35\textwidth}
\vspace{-5mm}
\centering
\includegraphics[width=0.99\linewidth]{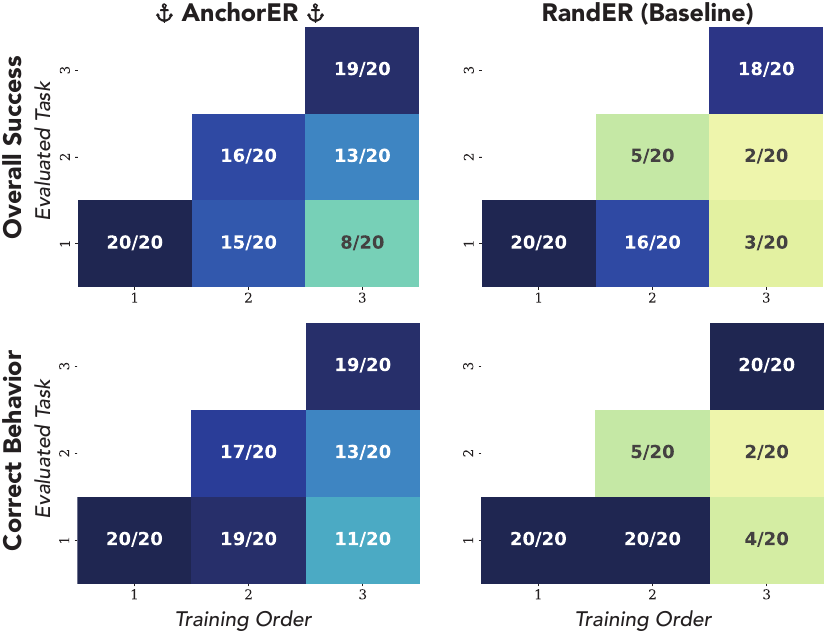}
\vspace{-2mm}
\caption{\textbf{Real World Results by Correct Behavior.} Some failures for both \methodname{} and \basemethod{} are due to lack of precision errors, not errors of behavior.}
\label{fig:app_realworldaltresults}
\vspace{-4mm}
\end{wrapfigure}

\cref{fig:realrobot} reports the overall successes of each policy on each behavior. We can further break down the performance of \methodname{} and \basemethod{} into failures of \textit{execution} and failures of \textit{overall strategy} in \cref{fig:app_realworldaltresults}. The training order is \textbf{Counterclockwise}, \textbf{Clockwise}, \textbf{Lift}. 

\textbf{Successes.} Success with the trained diffusion policy would make contact with the correct gripper(s) to move the lid in the desired direction. In some cases, because we only employed RGB camera feeds, the robot would miss contact initially, but because we demonstrated a neutral pose recovery in the training data, the policy would always retry until correct. After learning \textbf{Clockwise} (Task 2), the policy for both \methodname{} and \basemethod{} on the \textbf{Counterclockwise} task occasionally exhibited \textbf{Clockwise} behavior briefly before executing the correct \textbf{Counterclockwise} behavior. This happened at the high conflict initial jar state, before any rotation occurs. After the initial correct rotation direction, the policy across all methods appear more confident and generally finish executing the task. Therefore, the performance differences between \methodname{} and \basemethod{} represent the ability to distinguish these high action conflict states, a data region that Memory Anchor selection process concentrates (\cref{fig:app_anchorsvisual}).

\textbf{Loss of Plasticity.} When the policy lost plasticity in \basemethod{} while learning \textbf{Clockwise} (Task 2), it exhibited the behavior for \textbf{Counterclockwise} (Task 1) on the \textbf{Clockwise} (Task 2) jar consistently, leading to 15/20 failures (\cref{fig:app_realworldaltresults}). We did discover that by rotating the \textbf{Clockwise} jar 15-20 degrees in the right direction, the robot was able to finish the task correctly. This indicates that the behavior is learned but in competition around the high action conflict area (the neutral position), leading to failure. In contrast, using \methodname{} was able to learn \textbf{Clockwise} with only 3/20 rollouts showing the incorrect \textbf{Counterclockwise} behavior (\cref{fig:app_realworldaltresults}). 

\textbf{Catastrophic Forgetting.} When the policy with \basemethod{} forgot \textbf{Counterclockwise} while learning \textbf{Lift}, it grabbed the jar lid on the \textbf{Counterclockwise} jar and attempted to lift directly (reflecting the strategy used for \textbf{Lift}), which caused a failure as the lid was still locked onto the jar. The same problem occurred with the \textbf{Clockwise} jar. Similar catastrophic forgetting behavior occurred with \methodname{}, but at lower frequencies. 

\textbf{Other Failures.} A small proportion of failures in  \methodname{} and \basemethod{} are failures of execution caused by lack of precision. The main failure mode is the robot ``unrotating'' the lid when the wrong gripper makes contact while it moves in for the final lid grab.

\subsection{Practical Lessons and Failure Analysis: OpenJar}
\label{app:realrobot_laundry}
To set up our robot tasks, we began by training specialist single task policies and tuning the setup and data mixture until the specialist policies got nearly 100\% success rates, ensuring that performance reduction during continual learning was due to the sequential training and not factors that made the individual tasks difficult to learn. We list our findings below.

\begin{itemize}[leftmargin=*, itemsep=0.5em, topsep=0.2em, parsep=0pt, partopsep=0pt]
 \item Encoder strength and pretraining: On our U-Net diffusion policy, we initially used a ResNet-50 encoder. We discovered that the ResNet-50 was much more sensitive to the exact lighting conditions (including shadows) than a pretrained VIT encoder. Training the VIT from scratch with the policy yielded full failures. 
 \item Needing better lighting conditions: initial rounds of training and deployment used normal room lighting, but the dark grip tape on the gripper and the dark jar lid made it difficult to perceive contact between the gripper and lid. Adding additional light made this contact visible. 
 \item Incidental unrotation: when the robot finishes rotating the lid, it needs to grab the lid to remove. Initial data collection strategy did not focus on high precision during this segment and the trained policy ended up incidentally making gripper contact with the lid and reversed the rotation progress. To fix this, we deliberately biased the gripper towards the correct rotation during the final grasp, meaning that any incidental contact would not unrotate the lid. 
\end{itemize} 

\textbf{Buffer size sensitivity} In \basemethod{}, we were able to discover a boundary between stability and plasticity according to the buffer size, supporting some concurrent work \cite{zhu_can_2026}. At 1000 memories, \textbf{Clockwise} could be learned at the cost of \textbf{Counterclockwise}, and at 5000 memories (selected), \textbf{Clockwise} could no longer be learned. 1000 memories also led to forgetting on \methodname{}, but at 5000 memories, \methodname{} exhibited both stability and plasticity by learning \textbf{Clockwise} while retaining \textbf{Counterclockwise}. 

\textbf{Incidental Result: Difficulty of Task Homogeneity.} During our initial data collection attempt, a small lighting difference was observed between tasks. This led to very successful continual learning, even for \basemethod{}. When we made the lighting constant between tasks, the continual learning challenges were revealed and the benefits for \methodname{} appeared. The lighting differences made our task suite more heterogeneous, with high-level observational differences reducing the representational overlap. These results, although anecdotal, further support our mechanism for catastrophic forgetting as observational overlap with action disagreement. The difference between homogeneous and heterogeneous tasks is interesting and deserves exploration in future work.

\subsection{Real Robot Details: SweaterFold}
The second task suite presented in \S \ref{sec:realrobot} consists of three behaviors involving the same sweater on a bimanual robot setup. In \textbf{Fold Left}, the left arm grabs the left sleeve and folds it across the chest. In \textbf{Fold Right}, the right arm grabs the right sleeve and folds it across the chest. In \textbf{Fold Up}, both arms grab the lower part of the sweater and folds it towards the collar. The starting environment is identical across tasks, so language specifies the behavior. We use the same UMI setup on an ARX arm, using bimanual data collected using two UMI devices \cite{patel2026behaviorpromptingpolicydemonstrations}. The policy architecture is similar to \textbf{OpenJar}, with an additional CLIP-style language embedding to encode the task description. For all ER experiments, we used a buffer size of $125$ transitions. This is a lower number than \textbf{OpenJar} because there is only one decision point at the start of the trajectory.

\section{Implementation Details}
\label{app:details}
\subsection{Memory Anchor Implementation}
\label{app:details_anchors}
We detail the process of finding Memory Anchors  in Algorithm~\ref{alg:anchorer} and describe each step below.

\textbf{Step 1: Observation Overlap.} We use the observation-level latent space for the retrieval of new task data that overlapped with old task data. This latent space included proprioception, language embedding, and visual embeddings concatenated together. For each new task data point, we computed distances to the old data points in the latent spaces (line 3). We used percentile-level statistics on these distances (nearest neighbor, 1\%, 5\%, 10\%) to represent each new data point as a feature vector. We then used k-means to group the new data points into two clusters. We pick the cluster with the smallest mean distance as the observational overlap cluster (the output of step 1, line 5). We chose clustering instead of thresholding because different tasks could have very different levels of overlapping representations.


\begin{algorithm}[t]
\caption{Finding Memory Anchors}
\label{alg:anchorer}
\begin{algorithmic}[1]
\Require Past data $\mathcal{D}^{<n}=\bigcup_{i=1}^{n-1}\mathcal{D}^i$, new data $\mathcal{D}^n$, old policy $\pi_{\theta}^{n-1}$, latent encoder $\phi_{\theta}^{n-1}$, number of memory anchors to extract $M_A$
\Ensure Selected Memory Anchors $\mathcal{D}^{\text{Anchor}}$ for use in \methodname{}. 

\Statex \textbf{Step 1: Find observationally overlapping new data.}
\For{$(s,a)\in\mathcal{D}^n$}
    \State Compute distances from $\phi_{\theta}^{n-1}(s)$ to all old-task latents $\{\phi_{\theta}^{n-1}(s'):(s',a')\in\mathcal{D}^{<n}\}$
    \State Represent $(s,a)$ by the lower-tail distance statistics $q(s)=[\mathrm{NN},1\%,5\%,10\%]$
\EndFor
\State $\mathcal{D}_{\mathrm{SID}}^n \gets$ the KMeans cluster of $\{q(s):(s,a)\in\mathcal{D}^n\}$ with smaller mean distance

\Statex \textbf{Step 2: Find new data with high action disagreement.}
\State Define $d_{\pi}(s,a)=\|\hat a-a\|_2$, where $\tilde a=a+\alpha\epsilon$, $\epsilon\sim\mathcal{N}(0,I)$, and $\hat a\sim\pi_{\theta}^{n-1}(\cdot\mid s,\tilde a)$
\State Compute the mean $\mu$ and standard deviation $\sigma$ of $d_{\pi}(s,a)$ on held-out old-task data
\State $\mathcal{D}_{\mathrm{OD}}^n \gets \{(s,a)\in\mathcal{D}_{\mathrm{SID}}^n : d_{\pi}(s,a)>\mu+2\sigma\}$

\Statex \textbf{Step 3: Retrieve old data closest to the disagreeing new data.}
\For{$(s',a')\in\mathcal{D}^{<n}$}
    \State Score $(s',a')$ by its median kNN distance to $\mathcal{D}_{\mathrm{OD}}^n$ in the latent space of $\pi_{\theta}^{n-1}$
\EndFor
\State \Return $\mathcal{D}^{\mathrm{Anchor}}\gets$ the $M_{\mathrm{A}}$ old samples with the smallest scores
\end{algorithmic}
\end{algorithm}

\textbf{Step 2: Action Disagreement.} From the observational overlap data, we compute the action disagreement. Because we are using a diffusion policy, we frame action disagreement as a \textit{denoising} approach. From the action label $a$, we sample  noise $\epsilon$ and compute $\tilde{a} = a + \alpha \epsilon$ using an action scheduler. Then, using the action scheduler, we denoise $\hat{a} \sim \pi(\cdot | s, \tilde{a})$. This denoising paradigm is better than sampling another $a \sim \pi(\cdot | s)$ from pure noise in the case of action multimodality (line 6). The $\alpha \epsilon$ at small magnitudes will preserve the overall direction, allowing $\pi$ to respect the modality and prevent false positives in action disagreement. In contrast, if the policy fully disagrees with the action, it will pull the noised action in a different direction, leading to high disagreement. The magnitude of $\alpha$ is determined by the scheduler step, which is a hyperparameter. We set it to $20$ on a DDIM step schedule of $100$. We select the action disagreement data by computing a baseline disagreement mean $\mu$ and standard deviation $\sigma$ using the validation set of the old task data. Then, we select new task data that exceeds $\mu + 2 \sigma$ action disagreement (line 8), which represents a significant deviation from baseline action disagreement. We have no quota for this selection; for task suites that are highly heterogeneous (\S \ref{app:metrics_heterogeneous}), the entire observation overlapping set may be selected. For more homogeneous suites like LIBERO-Goal, this selection yields the critical decision points and excludes shared subtrajectories.   

\textbf{Step 3: Memory Anchor Retrieval.} We compute the $k$ nearest neighbors in the observation latent space from the old task data to the high action disagreement set from the new task data (line 10). The median of the $k$ nearest neighbors gives a score for each old data point. We can select Memory Anchors in increasing order of this score. 

\subsection{LIBERO Learning Policy Details}
The diffusion policy has a u-net backbone with 4 encoding and 4 decoding layers and a ResNet-18 image encoder based on a reference diffusion policy implementation from \cite{chi_diffusion_2023}. It takes a 2-step history stack of observations and predicts a chunk of 16 actions. It is trained with the standard diffusion policy objective and during inference, we use the noise predictor with the DDIM noise scheduler to craft the action chunk prediction. We use the Adam optimizer with a learning rate of 1e-4 and a cosine scheduler with 5\% warmup, reset per task. With LIBERO, we train each task for 50 epochs with batch size of 32. 

\subsection{LIBERO Experiment Details}
For all reported LIBERO results in the paper, we evaluate success rates using the $50$ environment reset configurations supplied by the benchmark \cite{liu_libero_2023}. For the continual learning results, we further report performances across three task permutations, also provided by the benchmarks \cite{liu_libero_2023}. The error bars in \cref{fig:mainresults} Left are therefore the NBTs for each task across all permutations, yielding $10 \times 3 = 30$ individual data points. For more details on the metrics, refer to \S \ref{app:details_metrics}. 

For results reported in \cref{fig:exp1} and \cref{fig:mainresults} Left, we used a total ER buffer of $1000$ consistent with the method in the reference ER implementation in the benchmark \cite{liu_libero_2023}. For the Anchor Addition experiment in \cref{fig:mainresults}, we chose an alternative ER allocation of $1\%$ past task data shown in other ER experiments on robots \cite{liu2026pretrained, zhu_can_2026}, which allowed an expanding buffer with increasing past tasks and increasing space to accommodate both Memory Anchors and a diverse representation of past task states.

\subsection{Continual Learning Metrics Details}
\label{app:details_metrics}

From the Libero benchmark \cite{liu_libero_2023}, there are two other continual learning metrics that reflect overall performance and new policy learning ability. Because the Negative Backward Transfer (NBT) fully covered the metric for forgetting, we chose to focus on NBT in the main paper. In this supplementary section, we show all three metrics for our experiments. To describe the three metrics, we define $c_{i, k}$ as the success rate on task $i$ after the model is trained up to task $k$. 

The main metric used in the paper, Negative Backward Transfer (NBT), measures the forgetting of a task as later tasks are trained. Each task $i < N$ has an NBT measurement defined as 

$$NBT_i = \frac{1}{N - i} \sum_{j = i +1}^N c_{i, i} - c_{i, j}$$

The reported metric is the average across all task $NBT_i$. The higher the metric, the worse the forgetting.

The Area Under Learning Curve (AUC) \cite{liu_libero_2023} is the normalized area under the success curve \cite{liu_libero_2023}, ranging from 0 to 1. Each task $i$ has a computed AUC defined as follows:

$$AUC_i = \frac{1}{N - i} \sum_{j = i}^N c_{i, j}$$

the reported metric is the average across all task $AUC_i$. Because it includes both the initial learning process and the later task retention, the AUC is a holistic measurement. 

Finally, the Forward Transfer (FWT) \cite{liu_libero_2023} is the performance of task $i$ after the initial training $FWT_i = c_{i, i}$, and the reported metric is the average of $FWT_i$. It measures the \textit{plasticity} of the policy to new information.

\begin{wrapfigure}{r}{0.35\textwidth}
\vspace{-5mm}
\centering
\includegraphics[width=0.99\linewidth]{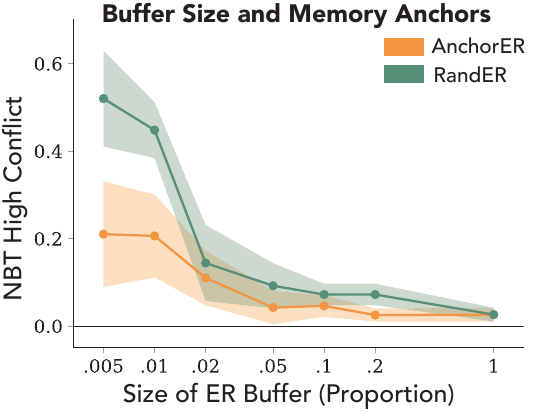}
\vspace{-2mm}
\caption{\textbf{Forgetting with Smaller Buffer Sizes.} As the available buffer size decreases, data retrieval through \methodname{} becomes more important. }
\label{fig:sweep}
\vspace{-5mm}
\end{wrapfigure}

\textbf{NBT On Task Conflict}. In our results, we discovered that a small proportion of task-task interactions contributed the majority of forgetting. To surface this effect, we propose a modified NBT that we use for \cref{fig:mainresults} Right (adding Anchor Memories) and \cref{fig:sweep}. We define a \textit{task-task interaction} between task $i$ and task $j$ as the segment $(c_{i, j-1}, c_{i, j})$. The delta $\Delta_{i, j} = c_{i, j-1} - c_{i, j}$ is how the performance on $i$ changes after being trained on task $j$. If $\Delta_{i, j} > 0$, then task $j$ causes task $i$ to decrease performance. The $NBT_i$ is the integration of $\Delta_{i, j}$ and reflects an aggregate metric of forgetting, while the $\Delta_{i, j}$ is a precise metric that measures task-task relationships. Each sequence of $n$ tasks has $n(n-1)/2$ such $\Delta_{i, j}$. To create our NBT Task Conflict metric, we use base ER runs to compute these $\Delta_{i, j}$ across three task training orders. Then, we pick the top $n$ tuples of $(i, j)$ and compute $\Delta_{i, j}$ across \methodname{} and \basemethod{} performances, which are graphed in \cref{fig:mainresults} Right. For our analysis, we use $n = 6$. For adherence to standard metrics, we also show the average NBT in Table \ref{tab:libero_results_addanchor}.  

\subsection{Memory Anchor Concentration Metric Details}

The Memory Anchor Concentration metric presented in \S \ref{sec:baselines} measures the presence of Memory Anchors in the buffer selected by each of the methods. It is computed retroactively and needs only the selected ER buffer and checkpoint for each task. First, the Memory Anchors are computed from the model checkpoints of each respective method. Then, we look at the buffer curated by the method and find the top 40\% sampled during training. The isolation of the top 40\% does not matter for non-MIR methods, as the other buffer curation approaches (including \methodname{}) will sample from the ER buffer randomly after curating them. We include the top 40\% condition because MIR selects counterexamples from an ER buffer based on the new task sample. 

The Anchor Concentration metric is the proportion of top 10\% Memory Anchors within this top 40\% buffer sample. The higher the proportion, the more Memory Anchors are being included. As seen in Table \ref{tab:baselines}, a randomly selected buffer already has concentration 0.1. Not shown, but in our subtraction experiment (\S \ref{sec:removeanchors}), the Concentration metric would show $0.0$ at the 10\% removal condition, as the 10\% removal would exclude all of the top 10\% Memory Anchors from the entire buffer.

\subsection{VLA Experiment Details}
\label{app:vla_details}

We use the OpenPi implementation of $\pi_{0.5}$ and all default full-model finetuning hyperparameters except for the number of steps per task, which we set to $5000$. We verified that $5000$ steps was enough to reach full convergence (\S \ref{app:vla}). For the latent space, we used the prefix embedding before the LLM backbone. We still used the action denoising step to compute action disagreement, but we applied the appropriate modifications for a flow-based model. Finally, to respect the original $\pi_{0.5}$ setup that does not keep a validation set, we used thresholding (top 10\% action disagreement) instead of the $\mu + 2\sigma$ selection for the action disagreement set, which would have required a validation set to compute $\mu$ and $\sigma$.

\section{Additional Simulation Results \& Interpretations}
\label{app:metrics}

\begin{table*}[t]
\centering
\small
\setlength{\tabcolsep}{6pt}
\resizebox{\textwidth}{!}{%
\begin{tabular}{lcccccc}
\toprule
\textbf{Access to Memory Anchors} &
\textbf{FWT($\uparrow$)} &
\textbf{NBT($\downarrow$)} &
\textbf{AUC($\uparrow$)} &
\textbf{FWT($\uparrow$)} &
\textbf{NBT($\downarrow$)} &
\textbf{AUC($\uparrow$)} \\
\midrule

& \multicolumn{3}{c}{\textsc{LIBERO-Long}}
& \multicolumn{3}{c}{\textsc{LIBERO-Spatial}} \\
\cmidrule(lr){2-4}
\cmidrule(lr){5-7}

100\% (\basemethod{})    & 0.87 $\pm$ 0.02 & 0.20 $\pm$ 0.02 & 0.73 $\pm$ 0.03 
    & 0.90 $\pm$ 0.02 & 0.05 $\pm$ 0.01 & 0.87 $\pm$ 0.01 \\
99\% (\basemethod{}-1\%)    & 0.86 $\pm$ 0.02 & 0.17 $\pm$ 0.02 & 0.74 $\pm$ 0.02
    & 0.90 $\pm$ 0.02 & 0.07 $\pm$ 0.02 & 0.85 $\pm$ 0.02 \\
95\% (\basemethod{}-5\%)     & 0.87 $\pm$ 0.02 & 0.28 $\pm$ 0.03 & 0.67 $\pm$ 0.03
    & 0.88 $\pm$ 0.02  & 0.14 $\pm$ 0.03 & 0.78 $\pm$ 0.03 \\
90\% (\basemethod{}-10\%) & 0.86 $\pm$ 0.02 & 0.34 $\pm$ 0.04 & 0.63 $\pm$ 0.04 
    & 0.90 $\pm$ 0.01 & 0.33 $\pm$ 0.04 & 0.66 $\pm$ 0.04 \\
\midrule

& \multicolumn{3}{c}{\textsc{LIBERO-Object}}
& \multicolumn{3}{c}{\textsc{LIBERO-Goal}} \\
\cmidrule(lr){2-4}
\cmidrule(lr){5-7}

100\% (\basemethod{})    & 0.95 $\pm$ 0.01 & -0.01 $\pm$ 0.01 & 0.96 $\pm$ 0.00 
    & 0.93 $\pm$ 0.01 & 0.05 $\pm$ 0.01 & 0.90 $\pm$ 0.01 \\
99\% (\basemethod{}-1\%)    & 0.95 $\pm$ 0.01 & 0.00 $\pm$ 0.01 & 0.95 $\pm$ 0.01 
    & 0.93 $\pm$ 0.01 & 0.07 $\pm$ 0.01 & 0.87 $\pm$ 0.02 \\
95\% (\basemethod{}-5\%)     & 0.96 $\pm$ 0.01 & 0.09 $\pm$ 0.03 & 0.89 $\pm$ 0.02
    & 0.93 $\pm$ 0.01 & 0.20 $\pm$ 0.03 & 0.79 $\pm$ 0.02 \\
90\% (\basemethod{}-10\%) & 0.95 $\pm$ 0.01 & 0.22 $\pm$ 0.05 & 0.79 $\pm$ 0.03
    & 0.92 $\pm$ 0.01 & 0.40 $\pm$ 0.05 & 0.64 $\pm$ 0.04 \\
\bottomrule
\end{tabular}
}
\caption{\textbf{Additional Metrics for Reducing Memory Anchor Access (\S \ref{sec:removeanchors})}. Reducing access to Memory Anchors reduces the ability to preserve past tasks but not the ability to learn new tasks. All numbers are from an ER buffer size of $1000$ sampled without replacement from the reduced past task dataset.}
\label{tab:libero_results}
\end{table*}

\begin{table*}[t]
\centering
\small
\setlength{\tabcolsep}{6pt}
\resizebox{\textwidth}{!}{%
\begin{tabular}{lcccccc}
\toprule
\textbf{Method} &
\textbf{FWT($\uparrow$)} &
\textbf{NBT($\downarrow$)} &
\textbf{AUC($\uparrow$)} &
\textbf{FWT($\uparrow$)} &
\textbf{NBT($\downarrow$)} &
\textbf{AUC($\uparrow$)} \\
\midrule

& \multicolumn{3}{c}{\textsc{LIBERO-Long}}
& \multicolumn{3}{c}{\textsc{LIBERO-Spatial}} \\
\cmidrule(lr){2-4}
\cmidrule(lr){5-7}

\basemethod{}
& 0.85 $\pm$ 0.02
&\textbf{ 0.25 $\pm$ 0.02}
& 0.67 $\pm$ 0.03
& 0.88 $\pm$ 0.01
& 0.18 $\pm$ 0.02
& 0.75 $\pm$ 0.01 \\

\methodname{} (10\%)
& 0.86 $\pm$ 0.02
& 0.26 $\pm$ 0.03
& 0.67 $\pm$ 0.03
& \textcolor{mygreen}{\textbf{0.90 $\pm$ 0.01}}
& \textcolor{mygreen}{\textbf{0.14 $\pm$ 0.01}}
& \textcolor{mygreen}{\textbf{0.79 $\pm$ 0.02}} \\

\methodname{} (20\%)
&\textbf{ 0.87 $\pm$ 0.02}
& 0.27 $\pm$ 0.03
&\textbf{ 0.68 $\pm$ 0.03}
& 0.88 $\pm$ 0.02
& 0.16 $\pm$ 0.02
& 0.76 $\pm$ 0.02 \\

\midrule

& \multicolumn{3}{c}{\textsc{LIBERO-Object}}
& \multicolumn{3}{c}{\textsc{LIBERO-Goal}} \\
\cmidrule(lr){2-4}
\cmidrule(lr){5-7}

\basemethod{}
& 0.96 $\pm$ 0.01
& 0.05 $\pm$ 0.01
& 0.92 $\pm$ 0.01
& 0.92 $\pm$ 0.01
& 0.18 $\pm$ 0.03
& 0.79 $\pm$ 0.03 \\

\methodname{} (10\%)
& 0.96 $\pm$ 0.01
& \textbf{0.03 $\pm$ 0.02}
& \textbf{0.93 $\pm$ 0.01}
& 0.93 $\pm$ 0.01
& \textcolor{mygreen}{\textbf{0.11 $\pm$ 0.02}}
& \textcolor{mygreen}{\textbf{0.85 $\pm$ 0.02}} \\

\methodname{} (20\%)
& 0.96 $\pm$ 0.01
& 0.05 $\pm$ 0.01
& 0.92 $\pm$ 0.01
& \textcolor{mygreen}{\textbf{0.94 $\pm$ 0.01}}
& 0.14 $\pm$ 0.02
& 0.84 $\pm$ 0.02 \\

\bottomrule
\end{tabular}
}
\caption{\textbf{Additional Metrics for Adding Memory Anchors (\S \ref{sec:addanchors})}. This table shows the \textit{average} CL metrics with a 1\% ER buffer (notice different buffer selection size than Table \ref{tab:libero_results}). Best results \textbf{bolded}, best results 1 SEM over baseline \textcolor{mygreen}{\textbf{in green.}} The homogeneous (Goal) task suite is affected more by additional Memory Anchors.}
\label{tab:libero_results_addanchor}
\end{table*}

\subsection{Additional Result: Memory Anchors on Smaller Buffer Sizes}
\label{sec:sweepanchors}

If Memory Anchors were indeed responsible for the varied forgetting across ER buffer samples (\S \ref{sec:rand_exp}), then smaller buffer sizes will decrease the likelihood of selecting them, making the effects of \methodname{} more visible. To test this, we sweep the ER buffer size from $0.5$\% to $100$\% (\cref{fig:sweep}). At higher buffer sizes, \basemethod{} and \methodname{} are indistinguishable because Memory Anchors are naturally present during uniform sampling. As buffer sizes become restricted, \methodname{} ensures adequate representation of Memory Anchors, while \basemethod{}'s forgetting increases sharply.

\subsection{Additional Result: Task-Task Relationships (\S \ref{sec:taskrelationships})}
\label{app:metrics_tasks}

\begin{wrapfigure}{r}{0.4\textwidth}
\vspace{-2mm}
\centering
\includegraphics[width=0.99\linewidth]{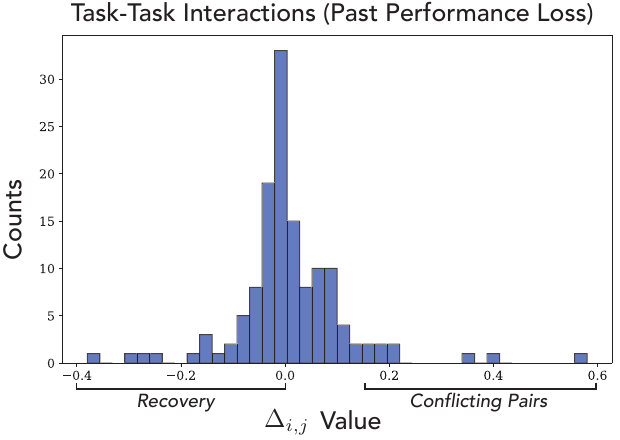}
\vspace{-2mm}
\caption{\textbf{Histogram of Task-Task Interactions ($\Delta_{i,j}$)} Most task-task interactions yield little forgetting, and a small proportion conflict heavily, leading to high forgetting. Many negative $\Delta_{i,j}$ instances exist even with an overall positive NBT because tasks will rebound after interacting with a conflicting task.}
\label{fig:app_histnbt}
\vspace{-6mm}
\end{wrapfigure}

As established in Section \ref{sec:taskrelationships}, it is important to look at individual task-task interactions, defined as how some task $T^i$ reacts when task $T^j$ is trained. We advocate for this perspective because the NBT metric collapses all of the interactions into a single number, while the task-task interactions have a high diversity of outcomes. NBT also unfairly penalizes high conflict task interactions earlier on in task training, because it uses the original task performance as the reference.

Plotting all of the interactions shows that some yield strong forgetting, while others only weakly, and still others even improve past task behavior (\cref{fig:app_histnbt}). Most full task $NBT_i$ is positive, meaning that the negative individual $\Delta_{i, j}$ (positive backward transfer) is the result of \textit{recovery} after a forgetting event. As established in past works \cite{liu2026pretrained}, forgetting past task behavior doesn't result in full destruction; rather, the past task is merely hidden and easily recoverable. Hence, after finishing training on the conflicting task, the affected task will rebound. All three examples in \cref{fig:taskint} show performance rebounding after the conflicting task stops training.

The task-task interactions are not only varied, they also contain notable outliers. As discussed in \S \ref{sec:taskrelationships}, a small proportion of interactions yield the most forgetting. Table \ref{tab:task_interactions} shows the top 5 highest conflict interactions in LIBERO-Goal. We can partition all of the $\Delta_{i, j}$ into the top 5 (shown above) and the bottom 75\footnote{Computed across three training orders. Does not add up to $3 \times 45$ because some task-task interactions are repeated across training orders}. The average bottom 75 $\Delta_{i, j}$ was $-0.0016$, and the average top $5$ was $0.207$. These numbers support our main paper claim: a small number of task-task conflicts drive much of the NBT score. The rest of the interactions are much smaller performance degradations and/or \textit{recovering} from previous conflicts.



\begin{table}[t]
\centering
\begin{tabularx}{\linewidth}{Xc}
\toprule
\textbf{Task Interaction} & \textbf{Mean $\Delta$} \\
\midrule
put the cream cheese in the bowl $\rightarrow$ put the bowl on the plate & 0.240 \\
put the bowl on top of the cabinet $\rightarrow$ put the wine bottle on top of the cabinet & 0.240 \\
put the wine bottle on the rack $\rightarrow$ turn on the stove & 0.180 \\
open the top drawer and put the bowl inside $\rightarrow$ put the wine bottle on top of the cabinet & 0.180 \\
turn on the stove $\rightarrow$ open the middle drawer of the cabinet & 0.120 \\
\bottomrule
\end{tabularx}
\vspace{2mm}
\caption{\textbf{Top five task-task conflicts in LIBERO-Goal.} This table shows the task-task interactions that have the highest forgetting (\S \ref{app:details_metrics}).}
\label{tab:task_interactions}
\end{table}



\subsection{Additional Result: On \naivemethod{} and \basemethod{} (Section \ref{sec:rand_exp})}
\label{app:metrics_base}
\begin{figure}[t]
  \centering
  \includegraphics[width=0.95\textwidth]{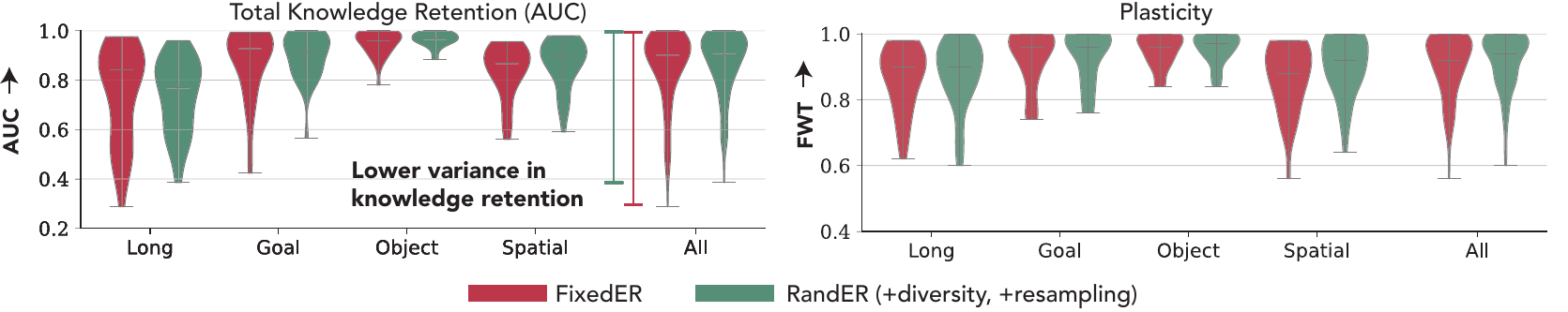}
  \caption{\textbf{Other metrics for comparing \naivemethod{} and \basemethod{}}. In addition to NBT shown in \cref{fig:exp1}, we show the AUC and FWT here. 
  }
  \label{fig:app_exp1}
    \vspace{-5mm}
\end{figure}

\cref{fig:app_exp1} shows the other metrics of the experiment in \S \ref{sec:rand_exp}. Consistent with the observation of the NBT, the overall performance AUC also reflects a reduced variance in $AUC_i$ performance. The forward transfer shows less of an impact with the ER sampling choice, reflecting that the sampling impacts more task retention (stability) than task acquisition (plasticity).

\subsection{Interpretation: Memory Anchors on Heterogeneous Tasks}
Our method of finding Memory Anchors through representation overlap and action disagreement is intuitive for homogeneous tasks, where tasks share common observations but differ in critical decision areas. Homogeneous tasks represent many common continual learning scenarios in the real world, including our real world task suite (\S \ref{sec:realrobot}). However, three of the LIBERO suites employed \textit{heterogeneous} tasks, which are suites with changed environments and/or object arrangements and identities between tasks, leading to distinctive observation spaces for each task. In highly heterogeneous task suites, removing Memory Anchors still increases forgetting and adding Memory Anchors still affects high conflict task-task interactions (\S \ref{sec:removeanchors}, \ref{sec:addanchors}). 

On heterogeneous tasks, the observation latent overlap will extract the closest possible points in the new data to the old task data. Often, these are at the very start of the trajectory because the robot has the same reset in the LIBERO suites. In highly heterogeneous suites, the action disagreement selection may include all of the new data points isolated by the representation overlap, as the policy fails to generalize to the new environment. Therefore, in highly heterogeneous suites, finding Memory Anchors becomes a filtered nearest-neighbor retrieval, which is a valid heuristic for selecting the most important past data to keep, as reflected by the impact of these extracted Memory Anchors on task performance. 

\label{app:metrics_heterogeneous}

\subsection{Additional Result: All Metrics for Memory Anchor Removal (Section \ref{sec:removeanchors})}
\label{app:metrics_memanchors}
The results in \cref{fig:mainresults} Left showed the average NBT for each task suite. In Table \ref{tab:libero_results}, we include the other two metrics in the experiment discussed in \S \ref{sec:removeanchors}. Although the ER buffer size remains constant, NBT rises and AUC drops significantly as the memory anchors become less available. Forward transfer (FWT) remains unaffected, which supports once again that the ER buffer mostly determines the stability of past policies, not the plasticity to future policies. However, this observation is not fully consistent between LIBERO and the real robot tasks. For a counterexample, refer to the real robot results in \S \ref{sec:realrobot}.

\subsection{Additional Result: All Metrics for Memory Anchor Enrichment (Section \ref{sec:addanchors})}
\label{app:metrics_anchorer}

The results in \cref{fig:mainresults} Right showed the NBT on the six most conflicting task-task pairs between \basemethod{} and \methodname{} (\S \ref{sec:addanchors}). We defined Task Conflict NBT (\S \ref{app:details_metrics}) using our findings from \S \ref{sec:taskrelationships} that a small number of task interactions contributed to most of the forgetting in a training sequence. However, to maintain consistency, we also report the three established metrics from this experiment in Table \ref{tab:libero_results_addanchor}. As mentioned in \S \ref{sec:addanchors}, the impact of an enriched ER buffer in \methodname{} is overall positive, with overall higher AUC and overall lower NBT between tasks.

These raw metrics also expose a critical difference in task suites discussed in \S \ref{sec:addanchors}. LIBERO-Goal is a homogeneous set of tasks with the same environment setup and objects between tasks. In this setup, \methodname{} is able to provide measured ($>1$ SEM) improvement across all three metrics. For the heterogeneous task suites (Long, Spatial, Object), the exact choice of ER matters less than the diversity present that covers all the prior task states, but \textit{removing} the computed Memory Anchors will still consistently reduce performance (\S \ref{sec:removeanchors}), an indication that the Memory Anchors are still serving a purpose. In these cases, the anchors present through random selection are sufficient (but necessary, as seen through the removal experiments), and heterogeneous tasks require more diversity in the ER buffer for wider state coverage. Adding more Memory Anchors trades this diversity for specificity. Homogeneous tasks benefit from this tradeoff, while heterogeneous tasks only benefit on the high-conflict task interactions.

\begin{wrapfigure}{r}{0.4\textwidth}
\vspace{-2mm}
\centering
\includegraphics[width=0.99\linewidth]{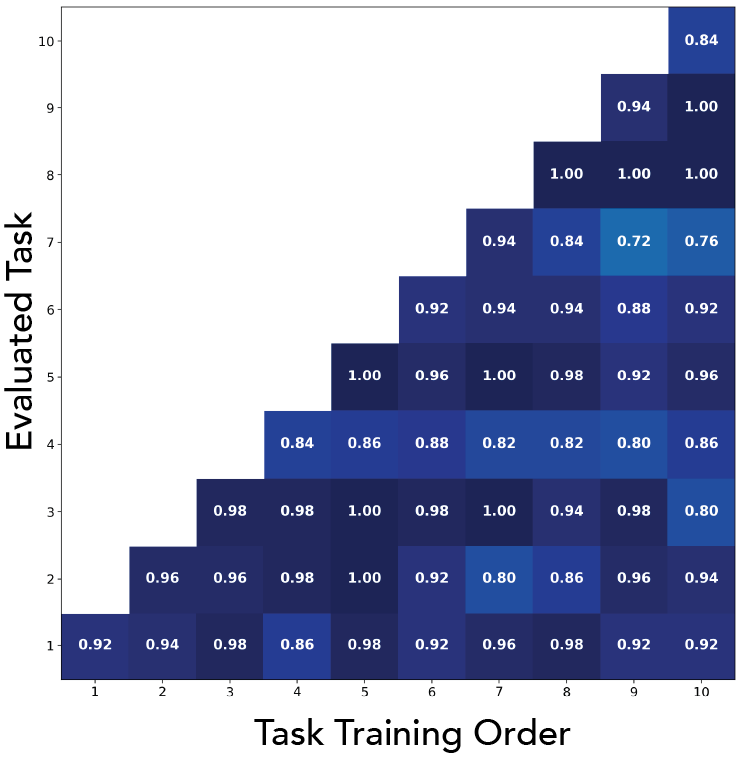}
\vspace{-2mm}
\caption{\textbf{$\pi_{0.5}$ on 1000 memories (20\%) buffer.} We verify that with a large ER buffer, $\pi_{0.5}$ achieves high task retention (NBT = 0.02 on LIBERO-Goal).}
\label{fig:app_vla}
\vspace{-6mm}
\end{wrapfigure}

\subsection{Additional Result: VLA Continual Learning Works at Large Buffers}
\label{app:vla}
Previous work that demonstrated continual learning performance on VLAs showed results with $\pi_0$. We conduct one of their experimental setups using the newer $\pi_{0.5}$: 1000 memories \textit{per task} (20\% of total past data) on LIBERO-Goal. We verify that this newer VLA also achieves high continual learning performance with average NBT $0.02$. We show a training order performance in Fig. \ref{fig:app_vla}. It is worth noting that the full finetuning with $\pi_{0.5}$ is able to achieve very low NBT while achieving higher average success rates ($0.93$ vs. $0.73$ reported in Table 1 of \cite{hu2026simple}). The reported results in \S \ref{sec:vla} focus on using $\pi_{0.5}$ on small buffer sizes (1\% and 1000 \textit{total} memories), and we discover that it exhibits more forgetting than from-scratch diffusion policy. We also show that, like diffusion policies, this VLA is also sensitive to Memory Anchors.



\end{document}